\documentclass[a4paper,fleqn]{cas-dc}

\usepackage{amsthm}
\usepackage{pifont}
\usepackage{balance}

\usepackage[numbers]{natbib}

\usepackage{cleveref}

\usepackage{amssymb}

\usepackage[export]{adjustbox}

\usepackage{algorithm}
\usepackage{algpseudocode}
\usepackage{amsmath}
\usepackage{xcolor}
\usepackage{setspace}
\usepackage{booktabs} 
\usepackage{multirow} 
\usepackage{makecell} 
\usepackage{pifont}

\usepackage{subcaption}
\usepackage{pdflscape}
\usepackage{enumitem}
\usepackage{xspace}

\def\tsc#1{\csdef{#1}{\textsc{\lowercase{#1}}\xspace}}
\tsc{WGM}
\tsc{QE}
\tsc{EP}
\tsc{PMS}
\tsc{BEC}
\tsc{DE}

\begin{document}

\let\WriteBookmarks\relax
\def\floatpagepagefraction{1}
\def\textpagefraction{.001}
\shorttitle{Task-Aware Scanning Parameter Configuration for Robotic Inspection}

\title[mode=title]{Task-Aware Scanning Parameter Configuration for Robotic Inspection Using Vision Language Embeddings and Hyperdimensional Computing}


\author[1]{Zhiling Chen}

\author[2]{David Gorsich}

\author[2]{Matthew P. Castanier}

\author[1]{Yang Zhang}

\author[1]{Jiong Tang}

\author[1]{Farhad Imani\cormark[1]}


\cortext[1]{Corresponding author}
\address[1]{School of Mechanical, Aerospace, and Manufacturing Engineering, University of Connecticut, Storrs, Connecticut, USA}
\address[2]{US Army DEVCOM Ground Vehicle Systems Center, Warren, Michigan, USA}

\cortext[2]{DISTRIBUTION STATEMENT A. Approved for public release; distribution is unlimited. OPSEC10423}


\begin{abstract}
Robotic laser profiling is widely used for dimensional verification and surface inspection, yet measurement fidelity is often dominated by sensor configuration rather than robot motion. Industrial profilers expose multiple coupled parameters, including sampling frequency, measurement range, exposure time, receiver dynamic range, and illumination, that are still tuned by trial-and-error; mismatches can cause saturation, clipping, or missing returns that cannot be recovered downstream. We formulate instruction-conditioned sensing parameter recommendation; given a pre-scan RGB observation and a natural-language inspection instruction, infer a discrete configuration over key parameters of a robot-mounted profiler. To benchmark this problem, we develop Instruct-Obs2Param, a real-world multimodal dataset linking inspection intents and multi-view pose and illumination variation across 16 objects to canonical parameter regimes. We then propose ScanHD, a hyperdimensional computing framework that binds instruction and observation into a task-aware code and performs parameter-wise associative reasoning with compact memories, matching discrete scanner regimes while yielding stable, interpretable, low-latency decisions. On Instruct-Obs2Param, ScanHD achieves 92.7\% average exact accuracy and 98.1\% average Win@1 accuracy across the five parameters, with strong cross-split generalization and low-latency inference suitable for deployment, outperforming rule-based heuristics, conventional multimodal models, and multimodal large language models. This work enables autonomous, instruction-conditioned sensing configuration from task intent and scene context, eliminating manual tuning and elevating sensor configuration from a static setting to an adaptive decision variable.
\end{abstract}


\begin{keywords}
 Robotic Scanning System \sep Laser Profile Scanner \sep Embodied Inspection 
 \sep Hyperdimensional Computing
\end{keywords}
\maketitle


\section{Introduction}
\label{sec:intro}

Quality inspection is a core capability in modern manufacturing, which safeguards dimensional accuracy and surface integrity, supports defect detection and metrology, and enables feedback to upstream processes for continuous improvement.
As manufacturing shifts toward higher product variety, faster product iteration, and tighter tolerances, inspection must be both accurate and adaptable across changing parts, fixtures, and environments \cite{oztemel2020literature}.
Consequently, industrial inspection is increasingly automated by combining robotic manipulators with high-resolution 2D/3D sensing, including machine vision and optical range sensing \cite{papavasileiou2025quality, rescsanski2025towards}.
Compared with manual gauging or fixed automation, robot-mounted sensing offers reconfigurability and repeatability, which allows an inspection cell to be repurposed across products and tasks with minimal retooling.

Among industrial 3D sensing modalities, laser-based profilers and line scanners are particularly attractive for measurement-driven inspection.
A laser profiler acquires dense 2D cross-sectional profiles at high frequency, and the profiler is swept across the surface by robot motion, forming a 3D point cloud suitable for dimensional verification, geometry reconstruction, and surface defect inspection.
Such robotic scanning systems have been widely investigated and deployed in manufacturing contexts, including high-value domains such as aerospace and precision machining \cite{guo2026modeling}.
Accordingly, considerable research has focused on where and how the sensor should move, including coverage path planning (CPP) \cite{dhiman2023multimodal}, viewpoint selection \cite{jiang2024fisherrf}, and cycle-time reduction under collision and accessibility constraints \cite{vutetakis2025active}.
These motion-centric methods have significantly advanced inspection automation, especially for complex freeform geometries and large-scale workpieces, by improving coverage completeness, reducing occlusions, and optimizing sequencing of viewpoints or stations.

However, robotic inspection performance is not determined by motion planning alone.
For range sensing in particular, measurement fidelity is fundamentally governed by how the sensor is configured at acquisition time, not just where it is positioned.
Modern industrial laser profilers expose a rich set of sensing parameters (e.g., sampling frequency, measurement range, exposure/integration time, receiver dynamic range, and emitted light intensity), each dictating a distinct aspect of the measurement process.
These parameters jointly determine the effective spatial resolution, field of view, signal-to-noise ratio, and susceptibility to failure modes such as saturation, clipping, blooming, and missing returns \cite{liu2023geometric}.
For example, in line-profiling systems the sampling frequency (together with robot feed rate) determines the along-path point spacing, i.e., $\Delta s \approx v / f_s$, directly trading cycle time against geometric detail.
Similarly, the lateral measurement range trades coverage against lateral sampling resolution because the sensor must distribute a finite number of sampling points over the selected field of view.
Exposure time, dynamic range, and laser intensity govern the received signal level and the robustness of the measurement to surface reflectivity, texture, incidence angle, and environmental conditions, mediating the risk of underexposure (dropouts) versus overexposure (saturation/clipping).
Errors introduced at this sensing stage are often irreversible; once a scan is corrupted by saturation, clipped geometry, or insufficient sampling density, downstream algorithms cannot recover information that was never measured.
This sensing-first reality makes parameter configuration a central decision problem in embodied inspection.

Despite its practical importance, sensing parameter configuration in industrial robotic inspection remains predominantly manual.
In practice, operators rely on experience-driven trial-and-error to adjust exposure, measurement range, and laser intensity when confronted with new materials, surface finishes, or inspection objectives. While modern industrial controllers provide built-in auto-tuning or auto-gain functions, these mechanisms operate solely to maximize signal strength. They are agnostic to the inspection intent, incapable of distinguishing between a task requiring global outline measurement (prioritizing range and speed) versus fine surface defect detection (prioritizing resolution and exposure stability), often resulting in reduced cycle times or missing features.
This reliance is not merely procedural but structural as sensing parameters directly determine the fidelity of the acquired measurements, yet their optimal values depend jointly on surface properties, task requirements, and environmental conditions that vary across production batches.
The challenge is further amplified in high-mix settings, where inspection tasks span coarse global coverage, fine local detail inspection, and metrology, each imposing fundamentally different sensing requirements.
Recent work has begun to formalize scan-quality evaluation and point-cloud quality assessment for robotic scanning \cite{maisano2023dimensional}, yet the upstream question of how to select sensing parameters, particularly in a task-aware way, remains underexplored in robotics and manufacturing research.
In fact, most CPP and viewpoint-planning pipelines assume a fixed sensing configuration, treating sensor parameters as static system constants \cite{chen2025pso}.
This assumption is restrictive as even an optimal trajectory can yield unusable data if exposure, range, or sampling regimes are mismatched to the object appearance and inspection intent.

A second practical driver is that inspection objectives are typically communicated through human-readable work instructions. In real inspection cells, engineers specify intent using task language such as ``scan the entire top surface,'' ``inspect the solder joints in detail,'' or ``measure the depth of the cavity.''
These instructions encode inspection scope (global vs.\ local), granularity (outline vs.\ detail), and measurement priorities that directly influence the most appropriate sensing regime.
In parallel, natural-language interfaces and large language models are rapidly entering manufacturing software ecosystems as decision-support tools \cite{naghavi2025multimodal, xu2025embodied}.
Yet, current instruction-following systems in robotics and multimodal learning primarily focus on high-level semantic understanding and action selection; the sensing pipeline is usually assumed fixed, and low-level sensing decisions are rarely modeled as first-class outputs.
Although inspection intent is both accessible and machine-readable, it is still not connected to choices about sensor configuration.

\begin{figure}
  \centering
  \includegraphics[width=0.49\textwidth]{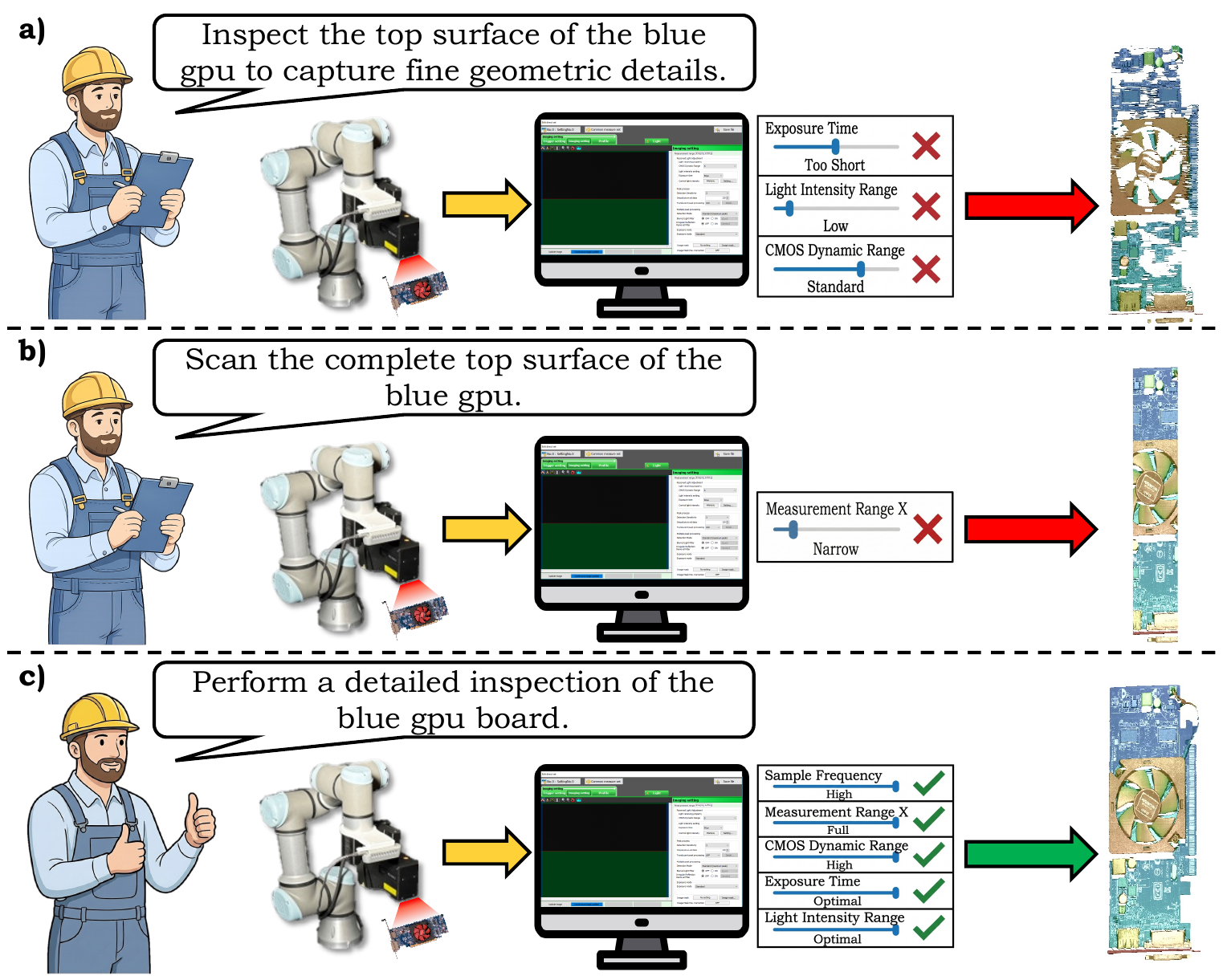}
  \caption{Instruction and observation dependent sensing parameter configuration in embodied inspection.
(a) A detail-oriented inspection instruction combined with insufficient exposure time leads to missing surface geometry.
(b) A global inspection instruction requires full measurement range, but an incorrect range setting results in clipped geometry.
(c) When sensing parameters are selected in accordance with both the inspection instruction and the observed object characteristics, the scan achieves sufficient fidelity for inspection.}
\vspace{-7mm}
  \label{introduction}
\end{figure}

As illustrated in Figure~\ref{introduction}, incorrect sensing configurations can arise from different sources.
In Figure~\ref{introduction}(a), a detail-oriented inspection instruction fails due to insufficient exposure, highlighting the importance of appearance-aware parameter selection.
In contrast, Figure~\ref{introduction}(b) shows that misunderstanding inspection intent can lead to inappropriate range settings and clipped geometry.
Only when both instruction intent and observable appearance are jointly considered does sensing configuration yield a usable scan, as shown in Figure~\ref{introduction}(c).

Motivated by these observations, we study instruction-conditioned sensing parameter recommendation for embodied industrial inspection. We view the laser profiler not as a static measurement device but as an adaptive sensing agent whose operating configuration should be selected according to both (i) the observed scene and object appearance and (ii) the task-level inspection intent expressed in natural language.
We assume a pre-scan visual observation (e.g., an RGB image from a wrist-mounted camera) and a natural-language inspection instruction are available before scanning begins.
The goal is to infer a discrete scanner configuration consisting of multiple controllable parameters, such as sampling frequency, measurement range, exposure time, receiver dynamic range, and light intensity.
The discrete nature is not an arbitrary modeling choice; industrial scanners often operate in a small number of stable, repeatable regimes used in practice for robustness and traceability.
Therefore, the learning problem is better posed as structured, multi-parameter classification under physical constraints, rather than unconstrained continuous regression.

A major barrier to progress is the lack of standardized datasets that jointly capture (i) real industrial appearance variation, (ii) task intent in natural language, and (iii) ground-truth scanner configurations that produce physically reliable measurements.
To enable systematic study, we create Instruct-Obs2Param, a real-world multimodal dataset collected on an industrially relevant robotic inspection platform using a UR3 manipulator and a Keyence LJ-X8200 laser profiler.
The dataset captures diverse objects, inspection tasks, viewpoints, and illumination conditions, reflecting variability encountered in practical inspection environments.
It is explicitly designed to probe whether models can produce stable parameter decisions under appearance changes, and whether they can adapt configuration based on instruction semantics when the same object must be scanned with different intent.

While the dataset enables benchmarking, the structure of instruction-conditioned sensing parameter reasoning poses challenges for conventional end-to-end learning.
The model must integrate heterogeneous signals (language intent and appearance cues), remain robust across illumination changes, and output interpretable, physically consistent decisions across repeated inspections.
In industrial settings, these properties are not merely desirable, they are often requirements for deployment, debugging, and traceability \cite{hoang2025hyperdimensional}.
To meet these requirements, we adopt a representation-first approach based on hyperdimensional computing (HDC) \cite{chen2025federated}.
HDC leverages high-dimensional distributed representations with algebraic operations (e.g., binding and bundling) that naturally support compositional fusion of multimodal inputs while providing intrinsic robustness to noise and spurious variations.
Equally important for inspection deployment, HDC inference reduces to simple similarity computations and associative retrieval, offering low latency and transparent decision pathways compared to opaque monolithic predictors.

Building on these insights, we propose ScanHD, an HDC-based framework for robust and interpretable sensing parameter recommendation in embodied inspection.
ScanHD binds visual observations and inspection instructions into symbolic high-dimensional representations and performs parameter-wise associative reasoning using compact memories.
This design yields stable decisions under appearance variation and physical constraints, and its lightweight inference is well suited for real-time integration into robotic inspection cells.
Experimental results on the Instruct-Obs2Param benchmark demonstrate that ScanHD achieves strong parameter prediction accuracy and robustness compared with rule-based heuristics, conventional multimodal learning methods, and multimodal large language model baselines, while maintaining low computational overhead. This paper makes the following contributions:
\begin{itemize}[noitemsep, topsep=0pt] 
    \item We formulate instruction-conditioned sensing parameter recommendation as a sensing-first embodied inspection problem, elevating scanner configuration from a fixed engineering setting to a first-class decision variable.
    \item We introduce Instruct-Obs2Param, a real-world dataset that links inspection intent, visual observations, and discrete laser scanning parameter configurations across objects and appearances.
    \item We propose ScanHD, a hyperdimensional computing framework that enables robust, interpretable parameter-wise reasoning for robotic scanning.
    \item We present comprehensive experiments on a real UR3-based laser scanning system, demonstrating improved accuracy, robustness, and deployment efficiency over strong baselines.
\end{itemize}

The remainder of this paper is organized as follows: 
Section~\ref{sec:related work} reviews related work in embodied inspection, robotic scanning systems, and hyperdimensional computing. 
Section~\ref{sec:method} introduces the Instruct-Obs2Param dataset, describing the hardware setup, task taxonomy, and data generation process. 
Section~\ref{research methodology} details the proposed research methodology, including the ScanHD framework and its hyperdimensional learning components. 
Section~\ref{ExperimentalResult} presents the experimental results and performance analysis. 
Finally, Section~\ref{Conclusion} concludes the paper and discusses future work.

\section{Overview of Related Work} \label{sec:related work}

\subsection{Robotic Inspection Planning and Sensing}

Embodied inspection systems integrate perception, physical embodiment, and task objectives to acquire informative observations from the environment. In contrast to navigation or manipulation tasks that primarily emphasize action feasibility or task completion, industrial inspection focuses on obtaining measurements with sufficient fidelity to support downstream analysis, such as defect detection \cite{chen2025multi}, dimensional verification \cite{liu2022coverage}, or quality assessment \cite{kim2015framework}. As a result, inspection performance is tightly coupled with how sensing is performed, rather than solely with where or when observations are collected.

Active perception has been extensively studied as a means to improve perception through action \cite{bajcsy1988active}, with a particular emphasis on viewpoint selection \cite{wang2023hierarchical}, next-best-view planning \cite{jin2023neu}, and coverage optimization \cite{egwuche2023machine}. These approaches enable agents to reason about which regions to observe and how to move sensors to reduce uncertainty or improve scene understanding. However, most existing active perception frameworks implicitly assume a fixed sensing pipeline, treating sensor parameters as static or externally specified throughout the perception–action loop.

This assumption is especially limiting in industrial inspection scenarios, where sensing parameters such as exposure time, measurement range, sampling frequency, and laser power settings have a direct and often nonlinear impact on observation quality. Different inspection tasks may tolerate vastly different trade-offs between spatial resolution, coverage, noise, and acquisition speed. Nevertheless, task semantics are rarely incorporated into the configuration of sensing parameters, leaving a critical gap between high-level inspection intent and low-level sensing behavior.

This gap motivates our work. Rather than treating sensors as passive collectors, we consider the laser profiler as a configurable sensing agent whose operating parameters fundamentally shape the quality and utility of the acquired data. To the best of our knowledge, existing embodied AI and active perception systems do not explicitly address the problem of predicting sensing parameters, such as exposure time, measurement range, illumination, and scanning frequency, jointly conditioned on visual observations and natural-language inspection instructions. This perspective highlights sensor parameter selection as a missing yet critical component of embodied intelligent inspection, motivating task-driven sensing adaptation.

\subsection{Industrial Robotic Scanning System}
Laser profilers play a central role in industrial inspection, providing high‐precision 2D surface measurements for defect detection \cite{gunatilake2020stereo}, dimensional verification \cite{torabi2021new}, and feature extraction \cite{wang2014multiscale}. Compared to RGB or depth cameras, 2D laser profilers offer micron‐level accuracy and robustness to lighting, making them indispensable in electronics, automotive, and semiconductor manufacturing. Modern systems, support adjustable exposure, measurement range, scanning frequency, and illumination settings. However, achieving high‐quality scans depends heavily on choosing the correct sensor parameters, which vary significantly across objects, materials, and inspection tasks.

In robotic inspection cells, laser profilers are typically mounted on manipulators for flexible scanning. Yet, despite automation in motion control, parameter configuration remains largely manual: operators must repeatedly adjust exposure, X/Z ranges, and lighting settings to avoid signal saturation, clipping, or low SNR. Such trial-and-error tuning is time-consuming, expertise-dependent, and difficult to scale across diverse geometries.
Although most industrial laser sensors provide built-in auto-gain or auto-optimization functions, these mechanisms are fundamentally task-agnostic: they blindly maximize signal intensity without considering the inspection objective. Consequently, they fail to reason about the trade-off between sensing cycle time (e.g., sampling frequency) and the geometric resolution required for defect detection or metrology, often resulting in suboptimal scans or unnecessary loss of efficiency.
Prior work in robotic perception has primarily focused on trajectory optimization or viewpoint planning for coverage, leaving the problem of sensor parameter selection almost unexplored.

Recent studies on robot learning and vision language models (VLMs) have enabled instruction-following and multimodal reasoning in manipulation tasks \cite{ai2025review}, but existing benchmarks overlook scenarios where the sensor itself is the primary agent and the goal is to determine how to sense rather than how to move. No standardized dataset evaluates the ability to interpret an observation, understand a natural-language inspection instruction, and infer the optimal sensing parameters.

This gap in existing robotic inspection research motivates the development of task-aware frameworks and datasets that explicitly study sensing parameter selection as a learning problem, enabling systematic investigation of how observation and task intent jointly inform effective sensing strategies in industrial environments.

\subsection{Instruction-Following Perception and Multimodal Robot Learning}

Recent advances in VLMs and vision-language-action (VLA) systems have enabled robots to interpret high-level semantic instructions and condition their perception and behavior accordingly \cite{shao2025large}. By integrating visual encoders with language-conditioned policies or planners, these approaches support instruction-guided object localization \cite{doveh2025teaching}, affordance detection \cite{engelbracht2025spotlight}, skill retrieval \cite{sarch2024vlm}, and manipulation planning \cite{feng2025reflective}. As a result, perception in robotic systems has increasingly shifted from task-agnostic processing toward task-conditioned interpretation driven by natural-language goals.

Despite their strong semantic reasoning capabilities, most instruction-following perception systems operate under the assumption of a fixed sensing pipeline. Language-conditioned perception typically modulates attention, selects regions of interest, or fuses multimodal representations, while leaving sensor operating parameters unchanged \cite{chen2025can}. Consequently, these models reason effectively about what to observe but do not address how the sensor should acquire observations. This limitation is particularly pronounced in industrial inspection settings, where sensing quality depends critically on task-specific configuration choices, such as adjusting exposure to capture fine surface details, narrowing measurement ranges for metrology, or modifying illumination to handle reflective materials.

Existing multimodal robot learning benchmarks further reflect this assumption by emphasizing manipulation, navigation, or object-centric reasoning tasks \cite{mees2022calvin, liu2023libero, li2023behavior}. Instruction-conditioned datasets primarily evaluate the ability to identify targets or select appropriate skills, rather than the ability to adapt sensing behavior. Even in neurosymbolic and compositional reasoning frameworks, the sensor is generally treated as a passive data source, with its configuration remaining decoupled from task intent.

These limitations motivate a complementary research direction in which instruction-following is extended beyond high-level task selection to include sensing configuration. In this setting, natural-language instructions and visual observations jointly inform how sensors should be configured to acquire task-appropriate data, bridging multimodal reasoning with actionable sensing adaptation.

\subsection{Hyperdimensional Computing}

HDC is a brain-inspired computing paradigm designed for efficient and robust learning, grounded in principles from theoretical neuroscience~\cite{kanerva2009hyperdimensional}. Owing to its high-dimensional distributed representations and simple algebraic operations, HDC has emerged as a promising alternative to deep learning for robotics applications that require reliability, interpretability, and computational efficiency.

A growing body of work has explored the use of HDC in robotic learning and control tasks. Early studies investigated HDC for navigation-related problems, including sequence-based localization, object recognition~\cite{neubert2019introduction}, and goal-oriented navigation~\cite{menon2022role}, demonstrating that hypervector representations can effectively encode sensor information and support reactive decision-making~\cite{neubert2017learning}. Subsequent efforts introduced extensions such as transfer learning and hybrid HDC–neural architectures to improve generalization and performance. While these approaches highlighted the potential of HDC in robotics, they were largely evaluated in simulation or offline settings and often relied on a limited number of discrete or low-dimensional sensor inputs. More recent work has begun to deploy HDC-based learning frameworks in real-world robotic systems, showing that hypervector representations can scale to continuous, high-dimensional sensory inputs such as LiDAR and support lightweight perception–action learning under real-world constraints~\cite{kwon2024brain}. These advances demonstrate that HDC can bridge the gap between computational efficiency and robust sensorimotor control, making it suitable for embodied robotic systems operating outside controlled laboratory environments.

Despite this progress, existing HDC-based robotics research has primarily focused on learning mappings from sensory observations to actions or control policies. The role of HDC as a reasoning substrate for higher-level sensing decisions, such as configuring sensor operating parameters in response to task intent and observed context, has received little attention.
 This gap suggests an opportunity to extend HDC beyond perception–action learning toward task-aware sensing adaptation, where discrete configuration choices must be inferred efficiently and robustly in industrial inspection scenarios.

\section{Instruct-Obs2Param Dataset} \label{sec:method}

\subsection{Hardware Setup} 

\begin{figure}[]
  \centering
  \includegraphics[width=\linewidth]{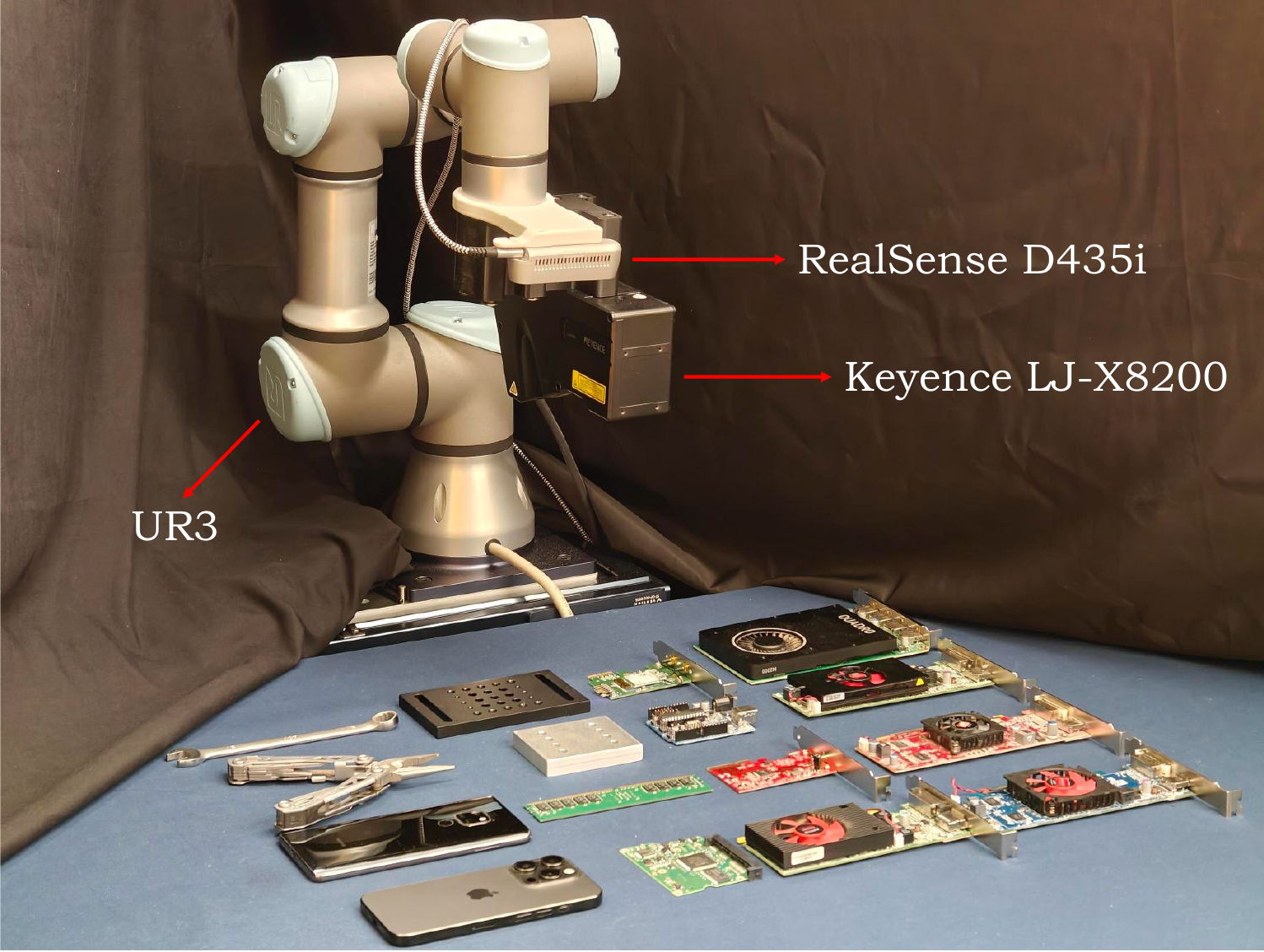}
  \caption{Hardware setup of the ScanBot system. A UR3 robotic arm is equipped with a Keyence LJ-X8200 laser profiler and an Intel RealSense D435i RGB-D camera mounted on the end-effector. A GoPro HERO8 captures third-person views from a fixed tripod. The entire setup operates within a black-curtained environment to ensure consistent and interference-free measurements.}
  \label{setup}
\end{figure}

As shown in Figure~\ref{setup}, the ScanHD experimental platform consists of a 6-DOF UR3 collaborative robotic arm equipped with a robot-mounted Keyence LJ-X8200 laser displacement sensor for industrial inspection. The laser profiler operates with a 405~nm blue laser and captures up to 3200 points per profile over an 80~mm measurement width, achieving 1~µm repeatability along the Z-axis and 3~µm resolution along the X-axis~\cite{keyence_ljx8200}. The laser profiler is interfaced with an LJX-8000A controller for data acquisition and parameter configuration. The laser scanner serves as the primary measurement device, while RGB observations of the target objects are acquired to support instruction-conditioned parameter reasoning in embodied inspection.

The experimental workspace is enclosed by black curtains to establish a controllable illumination environment. This setup enables us to systematically vary lighting conditions when capturing RGB observations, allowing us to assess the robustness of parameter recommendation under different visual appearances. 
While the laser profiler itself is largely insensitive to ambient illumination, the clarity and quality of visual observations play a critical role in embodied inspection, where sensor parameters are inferred based on scene appearance and task intent.
To account for this factor, we capture visual observations under three controlled lighting conditions—full light, side light, and dark—by actively adjusting external illumination using a controllable LED light panel.
These conditions are designed to emulate common real-world inspection scenarios, ranging from uniform illumination to directional lighting and low-light environments.
By evaluating the model across these settings, we explicitly examine whether the proposed approach can maintain consistent parameter recommendations despite changes in illumination, thereby validating its robustness to appearance variations encountered in real-world robotic inspection scenarios.

\subsection{Objects and Scanning Scenarios} 

\begin{figure}[]
  \centering
  \includegraphics[width=\linewidth]{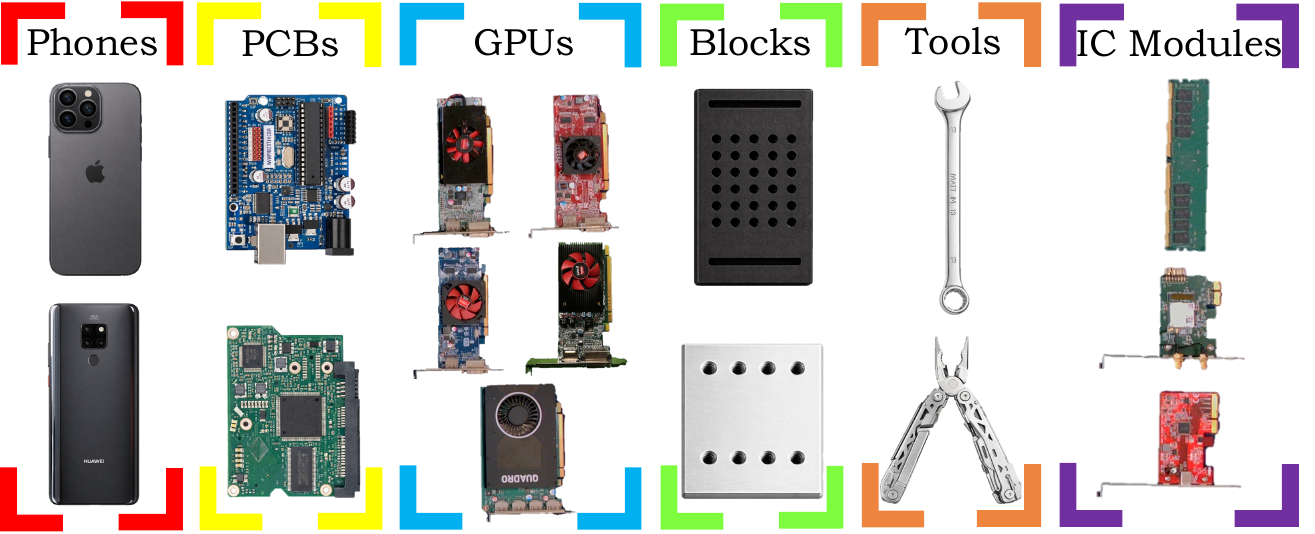}
  \caption{The dataset comprises 16 representative objects commonly encountered in robotic inspection, including consumer electronics (e.g., smartphones), printed circuit boards, GPU modules, mechanical tools, and calibration blocks. These objects exhibit diverse geometric scales, surface reflectivity, and structural complexity, posing varying challenges for scanning.}
  \vspace{-7mm}
  \label{objects}
\end{figure}

The Instruct-Obs2Param dataset includes 16 representative objects that reflect common robotic inspection and remanufacturing scenarios, spanning consumer electronics, PCBs, GPU modules, mechanical tools, and calibration artifacts, as illustrated in Figure~\ref{objects}. These objects are selected to cover a wide range of physical and visual properties, including variations in geometric scale, surface material, reflectivity, and structural complexity.

From a scanning perspective, the selected objects impose heterogeneous requirements on laser profiling. For instance, densely populated PCBs and GPU modules contain fine-grained components, edges, and cavities that demand high sampling resolution and precise exposure control, while larger and more uniform objects such as calibration blocks emphasize global surface coverage and measurement range. Metallic tools and reflective surfaces further introduce challenges related to signal saturation and dynamic range selection. By incorporating such diversity, the dataset captures realistic trade-offs encountered in practical inspection tasks.

Each object is associated with multiple scanning scenarios defined by different inspection intents and target regions. Rather than treating objects as isolated instances, instruct-Obs2Param emphasizes the interaction between object properties and task semantics. The same object may require substantially different scanning parameter configurations depending on whether the inspection focuses on global geometry, local outlines, fine structural details, or metrology-oriented measurements. This design encourages models to reason jointly over visual appearance, object characteristics, and instruction-specified intent when recommending scanning parameters.

\subsection{Parameter Space and Design Rationale}

\begin{figure*}[]
  \centering
  \includegraphics[width=\linewidth]{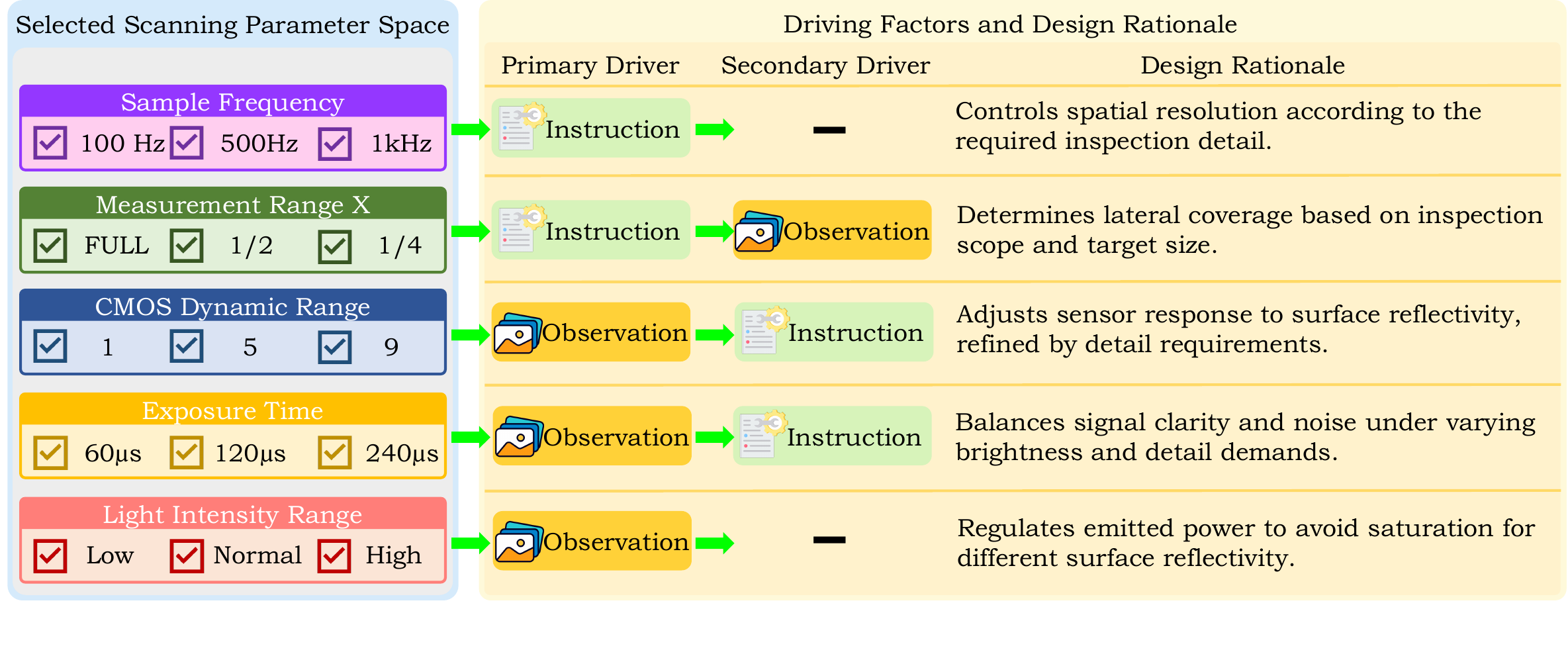}
  \caption{Five key scanning parameters are discretized into three representative options each, forming a compact and interpretable action space. For each parameter, we explicitly indicate its primary and secondary driving factors, distinguishing whether it is mainly determined by inspection intent (instruction) or by observation-level cues such as surface reflectivity and brightness.}
  \label{parameters}
\end{figure*}

Laser-based surface inspection systems expose a large number of configurable parameters, many of which are either hardware-specific or weakly coupled to inspection intent. In Instruct-Obs2Param, we focus on a compact yet expressive parameter space consisting of five key scanning parameters that are both practically controllable and strongly influence inspection outcomes. This selection is guided by two principles: (i) the parameters should capture the dominant trade-offs among coverage, resolution, and signal quality, and (ii) their optimal values should be meaningfully determined by inspection intent and observable object properties.

Specifically, we consider sampling frequency, measurement range along the X direction, CMOS dynamic range, exposure time, and light intensity range. As illustrated in Figure~\ref{parameters}, each parameter is discretized into three representative options, corresponding to stable operating regimes commonly adopted in inspection practice. Together, the parameters define the effective spatial resolution, field of view, and signal-to-noise characteristics of the scanning process. All parameter values are selected from representative operating configurations supported by the LJ-X8000 laser profiling system, and are chosen to span dominant sensing regimes rather than exhaustively sampling the continuous parameter space. Importantly, industrial laser profilers impose non-trivial coupling constraints across sensing parameters, 
such that not all combinations are physically realizable. 
Accordingly, the discrete action space in Instruct-Obs2Param is carefully designed to ensure that all parameter combinations correspond to physically valid and executable configurations on the LJ-X8000 system.

Sampling frequency is discretized into \{100\,Hz, 500\,Hz, 1\,kHz\}, directly controlling the density of acquired scan profiles and the level of geometric detail captured. Tasks that focus on coarse structural understanding favor lower sampling rates for efficiency, whereas fine-grained inspections demand higher frequencies to capture subtle geometric variations. Measurement range along the X direction is discretized into \{FULL, 1/2, 1/4\}, specifying whether the scan covers the entire object or a localized region of interest. This parameter is primarily driven by inspection coverage intent and is further constrained by observation-level factors such as the physical size of the target region. 

The remaining three parameters primarily regulate signal quality and robustness. CMOS dynamic range is discretized into \{1, 5, 9\} to accommodate surfaces with varying reflectivity characteristics, while exposure time is discretized into \{60\,µs, 120\,µs, 240\,µs\} to balance brightness, noise, and motion robustness under different inspection conditions. Although both parameters are largely influenced by observation-level cues such as material properties and illumination, inspection intent provides secondary refinement, as fine-detail tasks impose stricter requirements on signal clarity. Light intensity range is discretized into three bounded intervals (Low, Normal, High) and is predominantly observation-driven, regulating emitted power to prevent sensor saturation while maintaining sufficient signal strength across different surface reflectivity levels. By restricting the parameter space to these five dimensions with three discrete options each, Instruct-Obs2Param strikes a balance between physical realism and learning tractability.

\subsection{Task Taxonomy for Industrial Inspection}
\label{Task Taxonomy for Industrial Inspection}

Instruct-Obs2Param defines a task taxonomy to characterize the inspection intents supported in industrial quality inspection. Rather than treating instructions as unconstrained natural language prompts, we explicitly model inspection tasks at a semantic level, which serves as the foundation for instruction synthesis and parameter reasoning in the dataset.

We categorize industrial inspection tasks into two high-level groups: appearance inspection and dimensional and geometric metrology. Appearance inspection focuses on capturing visual and geometric characteristics of object surfaces, while dimensional and geometric metrology emphasizes precise measurement, alignment, and consistency across samples. 

Within appearance inspection, we further distinguish tasks based on inspection scope and granularity. Inspection scope is divided into global and local settings, corresponding to whether the scan targets the entire object or a specific region of interest. Inspection granularity is divided into outline and detail levels. Outline-oriented tasks aim to capture overall shape, contours, and structural boundaries, whereas detail-oriented tasks focus on fine geometric features, edges, and surface variations. Combining these dimensions yields four appearance inspection tasks: global outline, local outline, global detail, and local detail.

Dimensional and geometric metrology tasks target quantitative assessment rather than visual characterization. Instruct-Obs2Param includes metrology tasks that measure geometric properties such as depth, width, or spacing, as well as registration tasks that assess alignment and positional consistency across scans or samples. These tasks typically demand higher stability and consistency in scanning behavior, as they are sensitive to noise, misalignment, and parameter variation.

\subsection{Instruction Slot Representation and Language Realization}
\label{Instruction Slot}

Based on the task taxonomy for industrial inspection defined in the previous subsection, we introduce a structured instruction slot representation to encode inspection intent in a parameter-aware manner. Rather than treating instructions as free-form natural language, Instruct-Obs2Param represents each inspection intent using a fixed set of semantic slots that abstract task requirements independently of linguistic realization. This representation serves as an intermediate layer between high-level inspection objectives and low-level scanning parameter selection. Formally, an inspection intent is represented as a slot tuple
\begin{equation}
\mathbf{s} = \left( s_{\text{task}},\, s_{\text{coverage}},\, s_{\text{target}},\, s_{\text{detail}} \right),
\end{equation}
where $s_{\text{task}}$ denotes the inspection task category defined by the task taxonomy, $s_{\text{coverage}}$ specifies the spatial scope of scanning (e.g., global or local), $s_{\text{target}}$ identifies the region or component of interest, and $s_{\text{detail}}$ indicates the required inspection granularity. Each slot takes values from a finite and predefined vocabulary, enabling systematic enumeration of valid inspection intents.

The slot tuple $\mathbf{s}$ serves as a canonical representation of inspection intent and remains invariant across different linguistic realizations and visual observations. The instruction slot representation is explicitly designed to align with the discrete scanning parameter space introduced earlier. Different slots control complementary aspects of scanning behavior: the task and coverage slots govern inspection scope and field of view, the detail slot captures resolution requirements that influence sampling-related parameters, and the target slot provides semantic grounding for region-specific inspection.

By operating at the intent level, the slot-based representation decouples inspection objectives from surface-level variability in observations and language. Multiple observations of the same object, as well as multiple linguistic descriptions of the same inspection goal, may correspond to an identical slot configuration $\mathbf{s}$ and therefore share the same canonical scanning parameters.

\subsection{Data Evolution Flywheel}
\label{sec:data_flywheel}

\begin{figure*}[]
  \centering
  \includegraphics[width=\linewidth]{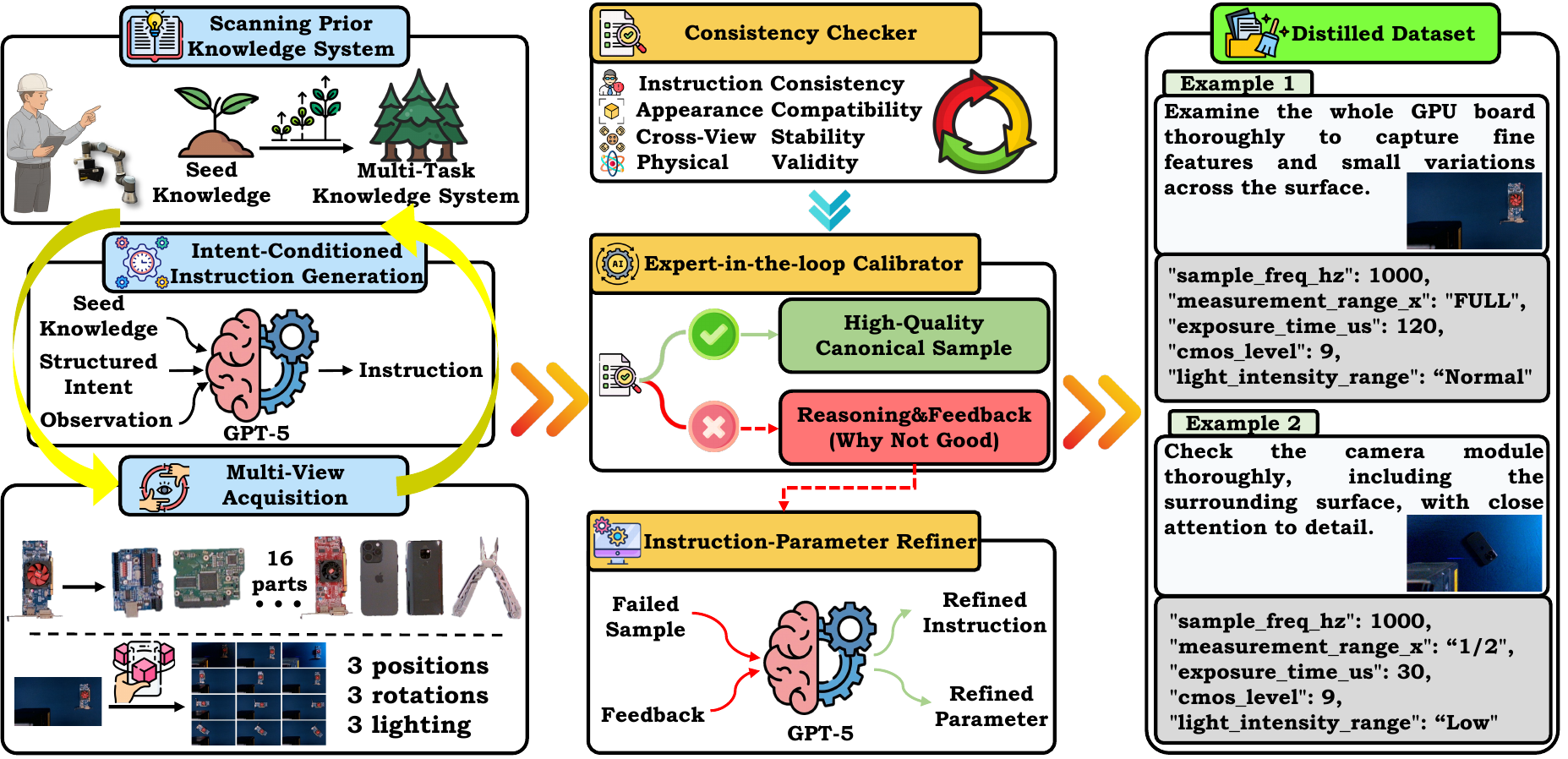}
  \caption{The Data Evolution Flywheel Framework: \textbf{Left:} intent-conditioned instruction instantiation based on structured inspection intent, multi-view observations, and scanning prior knowledge. \textbf{Middle:} consistency-driven checking, expert-in-the-loop calibration, and iterative instruction--parameter refinement. \textbf{Right:} representative instruction--observation--parameter instances distilled through the flywheel.}
  \label{flywheel}
\end{figure*}

Building upon the task taxonomy introduced in Section~\ref{Task Taxonomy for Industrial Inspection} and the instruction slot representation defined in Section~\ref{Instruction Slot}, we design a Data Evolution Flywheel for synthesizing high-quality instruction--observation--parameter triplets $(t,o,\vartheta)$ with an associated canonical intent label $\mathbf{s}$. The goal of this flywheel is to progressively align inspection intent, natural-language instructions, visual observations, and scanner parameters through an iterative generation, verification, and refinement process.

As illustrated in Fig.~\ref{flywheel}, the proposed framework consists of four tightly coupled stages:
(1) intent-conditioned instruction generation,
(2) consistency-driven checking,
(3) expert-in-the-loop calibration, and
(4) iterative refinement and dataset distillation.

\vspace{0.5em}
\noindent\textit{Intent-Conditioned Instruction Generation.}
Each inspection instance is first specified by a structured inspection intent $\mathbf{s}$, defined using the instruction slot representation introduced in Section~\ref{Instruction Slot}. Given a visual observation $o$ of the object and a scanning prior knowledge system $\mathcal{K}$, a strong generator $\mathcal{G}$ (e.g., GPT-5) is used to produce natural-language inspection instructions by sampling from a conditional language model:
\begin{equation}
t \sim \mathcal{G}\!\left(\cdot \,\middle|\, \mathbf{s}, o, \mathcal{K}\right),
\end{equation}
where $t$ denotes a linguistic realization of the inspection intent $\mathbf{s}$. Each instruction $t$ specifies what to inspect and at what level of detail, without exposing any scanner parameters or configuration details.

To promote robustness to appearance variations, each object is observed under multiple object poses and illumination conditions:
\begin{equation}
\mathcal{O} = \left\{ o^{(p,r,l)} \,\middle|\, p \in \mathcal{P},\; r \in \mathcal{R},\; l \in \mathcal{L} \right\},
\end{equation}
where $\mathcal{P}$ and $\mathcal{R}$ denote the sets of object positions and rotations, respectively, and $\mathcal{L}$ denotes the set of lighting conditions. All observations share the same inspection intent $\mathbf{s}$, ensuring that instruction generation is invariant to pose- and illumination-induced appearance changes.

\vspace{0.5em}
\noindent\textit{Consistency-Driven Checking.}
While intent-conditioned generation yields diverse instruction candidates, not all synthesized instances are suitable for embodied inspection. We therefore introduce a consistency checker $\mathcal{C}$ to verify the coherence of each generated instance across multiple dimensions:
\begin{equation}
\mathcal{C}(t, o, \mathbf{s}, \vartheta) \in \{0,1\}
\end{equation}
where $\vartheta$ denotes the associated scanner parameter configuration. Only samples satisfying all criteria are retained as valid candidates:
\begin{equation}
\mathcal{D}_{\text{cand}} = \left\{ (t,o,\vartheta,\mathbf{s}) \,\middle|\, \mathcal{C}(t,o,\mathbf{s},\vartheta)=1 \right\}
\end{equation}

Specifically, $\mathcal{C}$ evaluates:
(i) instruction consistency, ensuring semantic alignment between $t$ and the intent slots $\mathbf{s}$;
(ii) appearance compatibility, verifying that the instruction is plausible given the observed object appearance $o$;
(iii) cross-view stability, enforcing invariance of intent realization across observations $o \in \mathcal{O}$; and
(iv) physical validity, ensuring feasibility of the parameter configuration under hardware constraints. Instances that do not pass checking are preserved as hard negatives for subsequent diagnosis and refinement.

\vspace{0.5em}
\noindent\textit{Expert-in-the-Loop Calibration.}
Automated consistency checking alone is insufficient to capture nuanced inspection requirements. We therefore introduce an expert-in-the-loop calibrator that further evaluates the consistency-approved candidate set $\mathcal{D}_{\text{cand}}$ and partitions it as
\begin{equation}
\mathcal{D}_{\text{cand}} = \mathcal{D}_{\text{canon}} \cup \mathcal{D}_{\text{fail}} \qquad
\mathcal{D}_{\text{canon}} \cap \mathcal{D}_{\text{fail}} = \emptyset
\end{equation}
where $\mathcal{D}_{\text{canon}}$ contains high-quality canonical samples, and $\mathcal{D}_{\text{fail}}$ consists of samples requiring revision.

For each instance in $\mathcal{D}_{\text{fail}}$, structured expert feedback $\Delta_{\mathrm{fb}}$ is provided, identifying sources of inconsistency such as ambiguous detail emphasis, unstable parameter selection, or misalignment between instruction and appearance. This feedback is recorded and reused in the refinement stage.

\vspace{0.5em}
\noindent\textit{Iterative Refinement and Dataset Distillation.}
Samples in $\mathcal{D}_{\text{fail}}$, together with their associated expert feedback $\Delta_{\mathrm{fb}}$, are passed to an instruction--parameter refiner (e.g., GPT-5) that revises either the instruction or the parameter configuration:
\begin{equation}
(t', \vartheta') = \mathcal{G}_{\text{refine}}\!\left(t, o, \mathbf{s}, \vartheta, \Delta_{\mathrm{fb}}\right)
\end{equation}
where the refinement is explicitly constrained to preserve the original inspection intent $\mathbf{s}$.

We then re-evaluate refined samples with the same consistency checker, yielding a closed-loop evolution process. Let $\mathcal{D}^{\text{fail}}_{0}=\mathcal{D}_{\text{fail}}$ denote the initial pool to be refined. At iteration $k$, refinement produces
\begin{equation}
\widetilde{\mathcal{D}}_{k}
=
\left\{
(t', o, \vartheta', \mathbf{s})
\,\middle|\,
\begin{aligned}
&(t, o, \vartheta, \mathbf{s}) \in \mathcal{D}^{\text{fail}}_{k}, \\
&(t', \vartheta') =
\mathcal{G}_{\text{refine}}\!\left(
t, o, \mathbf{s}, \vartheta, \Delta_{\mathrm{fb}}
\right)
\end{aligned}
\right\}
\end{equation}
and the checker splits it into passing and remaining-failing subsets:
\begin{equation}
\mathcal{D}^{\mathrm{pass}}_{k+1}
=
\left\{
(t',o,\vartheta',\mathbf{s}) \in \widetilde{\mathcal{D}}_{k}
\;\middle|\;
\mathcal{C}(t',o,\mathbf{s},\vartheta')
\right\}
\end{equation}

\begin{equation}
\mathcal{D}^{\mathrm{fail}}_{k+1}
=
\widetilde{\mathcal{D}}_{k}
\setminus
\mathcal{D}^{\mathrm{pass}}_{k+1}
\end{equation}

Through repeated iterations, unstable generations are filtered out and canonical intent realizations are reinforced. After $K$ refinement rounds (or upon convergence), the final outcome of the flywheel is a distilled dataset
\begin{equation}
\mathcal{D}^{*} = \mathcal{D}_{\text{canon}} \cup \bigcup_{k=1}^{K} \mathcal{D}^{\text{pass}}_{k}
\end{equation}
where each instance consists of a natural-language instruction $t$, a visual observation $o$, a scanner parameter configuration $\vartheta$, and a canonical inspection intent $\mathbf{s}$.

\subsection{Data Statistics}
\label{sec:data_statistics}

\begin{figure*}[]
  \centering
  \includegraphics[width=\linewidth]{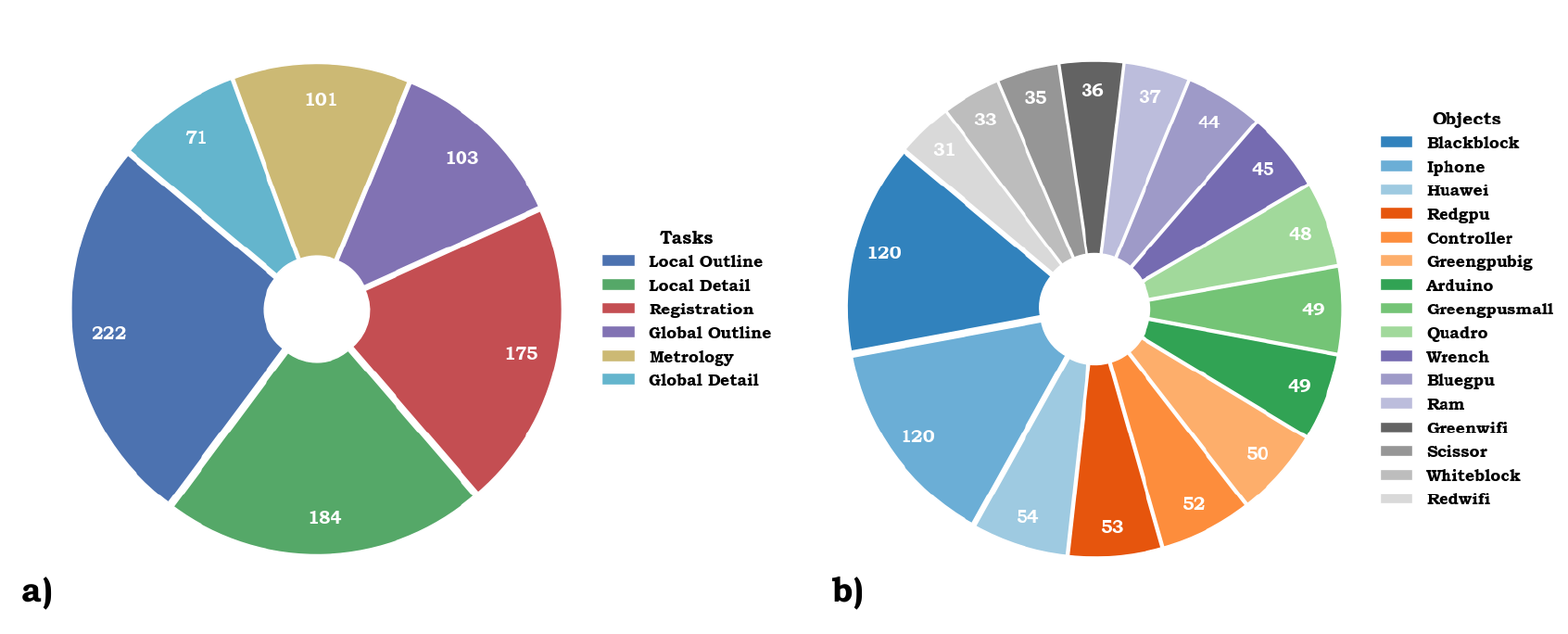}
  \caption{Dataset statistics of Instruct-Obs2Param.
(a) Distribution of synthesized instructions across different inspection task types. (b) Distribution of instructions across the 16 inspected objects.}
  \label{fig:data_stats}
\end{figure*}

The dataset consists of instruction--observation--parameter instances $(t,o,\vartheta)$ with an associated canonical intent label $\mathbf{s}$ (Sec.~\ref{sec:data_flywheel}). In practice, the dataset is organized around unique intent-conditioned instruction--parameter keys $(t,\vartheta,\mathbf{s})$, and for each such key, we collect multi-view RGB observations under multiple appearance conditions. Specifically, for each fixed $(t,\vartheta,\mathbf{s})$, we acquire an observation set
\begin{equation}
\begin{aligned}
\mathcal{O}
&= \left\{ o^{(p,r,l)} \mid (p,r,l) \in \mathcal{P} \times \mathcal{R} \times \mathcal{L} \right\}, \\
|\mathcal{P}| &= |\mathcal{R}| = |\mathcal{L}| = 3
\end{aligned}
\end{equation}

So that $|\mathcal{O}| = |\mathcal{P}|\,|\mathcal{R}|\,|\mathcal{L}| = 27$.
This design decouples instruction and parameter diversity from appearance variation, enabling systematic evaluation of robustness to viewpoint and illumination changes.

Figure~\ref{fig:data_stats}(a) shows the distribution of synthesized instructions across different inspection task types defined in the task taxonomy (Sec.~\ref{Task Taxonomy for Industrial Inspection}). The dataset covers a diverse set of inspection intents, including global and local appearance inspection, outline- and detail-level analysis, as well as metrology and registration tasks. No single task type dominates the dataset, indicating that instruction synthesis is not biased toward a narrow subset of inspection objectives. This balanced task-level distribution reflects the structured design of the task taxonomy and supports learning across multiple inspection behaviors.

Figure~\ref{fig:data_stats}(b) presents the distribution of instructions across the 16 inspected objects. The dataset spans consumer electronics, printed circuit boards, GPU modules, mechanical tools, and calibration artifacts, covering a wide range of geometric scales, surface reflectivity, and structural complexity. Instructions are relatively evenly distributed across objects, preventing over-representation of specific instances and enabling meaningful evaluation under object-level generalization settings.


\section{Research Methodology}\label{research methodology}

\subsection{Hyperdimensional Computing Primitives}

\begin{figure}[]
    \centering
    \includegraphics[width=0.5\textwidth]{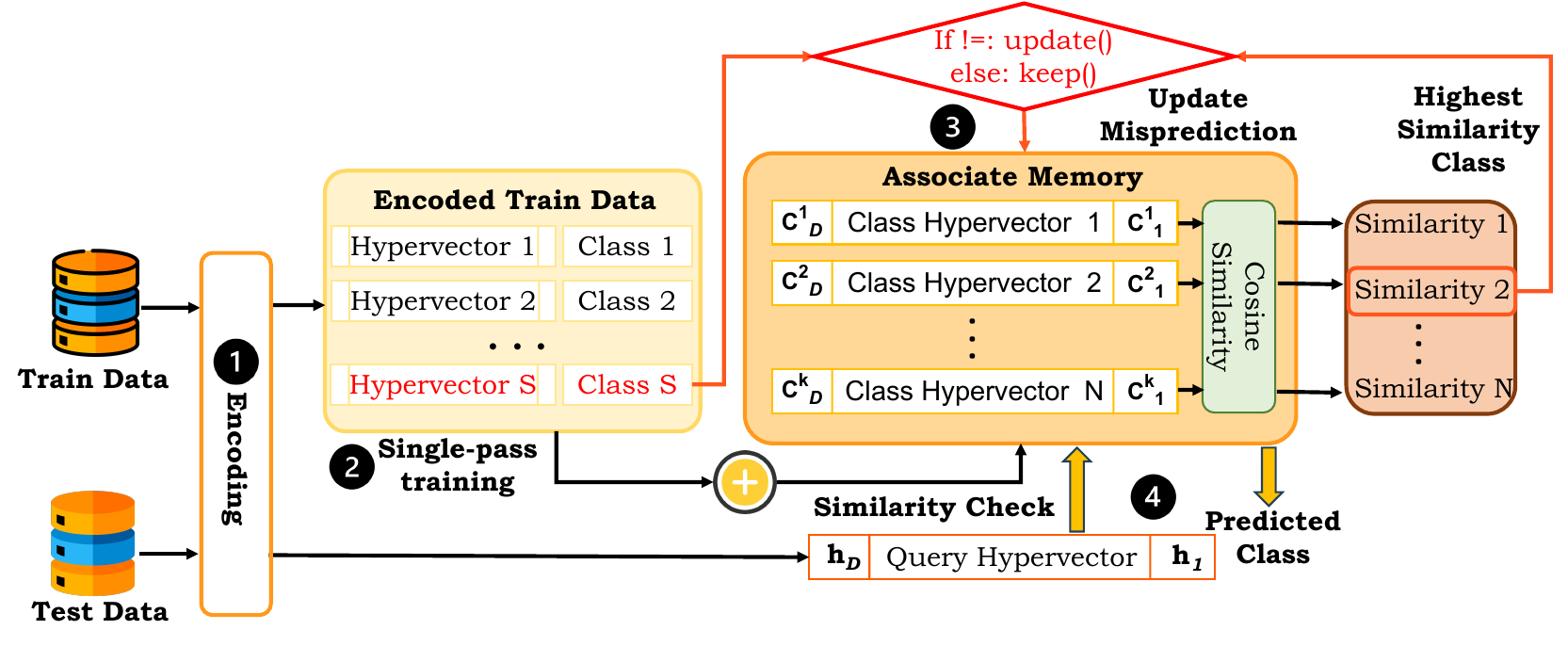}
    \caption{Overview of the HDC learning procedure. (1) Encode raw data into hypervectors. (2) Hypervectors from the same class are aggregated to create class hypervectors. (3) Update class hypervectors in response to misclassifications. (4) Compare query hypervectors to class hypervectors via similarity during inference.}
    \label{HDC}
\end{figure}

Figure~\ref{HDC} provides an overview of the HDC framework. The overall learning procedure consists of four stages: encoding, single-pass training, retraining, and inference.

As illustrated in Fig.~\ref{HDC}~\ding{182}, input data are projected into a high-dimensional bipolar (binary) space via a hyperdimensional encoder $\mathcal{E}(\cdot)$. For an input vector $\mathbf{x}\in\mathbb{R}^{d_x}$, we use the standard random-projection sign encoder
\begin{equation}
\mathbf{h} = \mathcal{E}(\mathbf{x}) = \operatorname{sgn}(\mathbf{E}\mathbf{x}) \in \{-1,+1\}^{D_h}
\end{equation}
where $\operatorname{sgn}(\cdot)$ denotes an element-wise sign function, and $\mathbf{E} \in \{-1,+1\}^{D_h \times d_x}$ is a fixed random bipolar projection matrix. This encoding process is built upon the fundamental operations of bundling and binding, which enable efficient combination and association of information in high-dimensional space.

The bundling operation aggregates multiple hypervectors through component-wise summation. Given three hypervectors $\mathbf{h}_1$, $\mathbf{h}_2$, and $\mathbf{h}_3$, the bundled representation is computed as
\begin{equation}
\mathbf{h}_b = \mathbf{h}_1 + \mathbf{h}_2 + \mathbf{h}_3
\end{equation}

The hypervector $\mathbf{h}_b$ remains more similar to each of its constituents than to unrelated hypervectors. Formally, defining the cosine similarity between two hypervectors as
\begin{equation}
\delta(\mathbf{h}_1, \mathbf{h}_2) =
\frac{\mathbf{h}_1^\top \mathbf{h}_2}{\|\mathbf{h}_1\|_2 \, \|\mathbf{h}_2\|_2}
\end{equation}

We typically observe $\delta(\mathbf{h}_b, \mathbf{h}_1) \gg \delta(\mathbf{h}_b, \mathbf{h}_4)\approx 0$ for an unrelated hypervector $\mathbf{h}_4$.
In contrast, the binding operation, implemented via the Hadamard (element-wise) product, associates multiple pieces of information and maps the resulting hypervector to a distinct and approximately orthogonal location in the hyperspace. For bipolar hypervectors, binding is defined as
\begin{equation}
\mathbf{h}_{\mathrm{bind}} = \mathbf{h}_1 \odot \mathbf{h}_2
\end{equation}
where $\odot$ denotes element-wise multiplication. After encoding, class hypervectors are constructed through a single-pass training procedure. As shown in Fig.~\ref{HDC}~\ding{183}, hypervectors corresponding to samples from the same class are bundled to form a unified class hypervector. The resulting class hypervectors are stored in an associative memory for subsequent inference.

To further refine the associative memory and improve classification accuracy, an additional retraining step is applied. As illustrated in Fig.~\ref{HDC}~\ding{184}, each encoded training hypervector $\mathbf{h}$ is compared against all class hypervectors in the associative memory using cosine similarity. The class with the highest similarity score is selected as the predicted label $\hat{y}$. If the prediction is correct, no update is performed. Otherwise, the class hypervectors are updated according to the following rules: $\mathbf{C}_{y} \leftarrow \mathbf{C}_{y} + \mathbf{h}$ and $\mathbf{C}_{\hat{y}} \leftarrow \mathbf{C}_{\hat{y}} - \mathbf{h}$, where $y$ denotes the true class label, $\hat{y}$ denotes the incorrectly predicted label, and $\mathbf{C}_{y}$ and $\mathbf{C}_{\hat{y}}$ are the corresponding class hypervectors. This error-driven update mechanism incrementally sharpens class boundaries in the hyperspace.

During inference, as shown in Fig.~\ref{HDC}~\ding{185}, an encoded sample is compared with all class hypervectors by computing cosine similarity, and the class with the highest similarity score is selected as the final prediction.

\subsection{Problem Definition and Design Objectives}
\label{sec:problem_definition}

We study the problem of instruction-conditioned sensing parameter recommendation for embodied industrial inspection.
Unlike conventional perception pipelines that treat sensing configurations as fixed, we consider the sensor itself as an adaptive agent whose operating parameters must be selected according to both the observed scene and the inspection intent.

\subsubsection{Problem Definition.}
Let $o \in \mathcal{X}$ denote a visual observation of a target object (e.g., an RGB image captured prior to scanning), and let $t \in \mathcal{T}$ denote a natural-language inspection instruction specifying what to inspect and at what level of detail.
The goal is to infer a discrete scanner parameter configuration
\begin{equation}
\vartheta = (\vartheta_1, \vartheta_2, \dots, \vartheta_P)
\qquad \vartheta_k \in \mathcal{Y}_k
\end{equation}
where $P$ is the number of controllable scanning parameters and $\mathcal{Y}_k$ is the discrete value set of the $k$-th parameter.
Each $\vartheta_k$ corresponds to a discrete operating choice of a controllable scanning parameter, such as sampling frequency, measurement range, exposure time, CMOS dynamic range, or illumination setting.
The inferred configuration $\vartheta$ should enable high-quality data acquisition that is consistent with the inspection intent expressed in $t$ and the physical properties observable in $o$.

This formulation differs from standard instruction-following perception and multimodal reasoning tasks.
Rather than predicting object labels, regions of interest, or action primitives, ScanHD focuses on predicting how the sensor should operate, transforming instruction-following from a high-level semantic understanding problem into a low-level sensing configuration problem.

\subsubsection{Design Objectives.}
To enable reliable robotic deployment, we design ScanHD around the following objectives:

\begin{itemize}
    \item \textbf{O1: Instruction-Conditioned Sensing Representation.}
    The representation should jointly encode the inspection instruction and visual observation, enabling task-aware sensing rather than task-agnostic perception. The representation must preserve semantic alignment with inspection intent while remaining robust to variations in object pose, viewpoint, and illumination.

    \item \textbf{O2: Parameter-Wise Decoupled Reasoning.}
    Scanning parameters influence sensing outcomes through physical mechanisms.
    The method should support independent yet coordinated reasoning over individual parameters, avoiding monolithic end-to-end regression that entangles unrelated decision factors.

    \item \textbf{O3: Lightweight and Interpretable Inference.}
    The resulting system should support efficient inference suitable for real-time robotic inspection and provide interpretable decision pathways, enabling analysis of how instruction and observation jointly influence each parameter recommendation.
\end{itemize}

These objectives motivate a symbolic, hyperdimensional approach in which instruction-conditioned observations are encoded into a structured representation that supports compositional reasoning and efficient associative inference.
The proposed ScanHD framework operationalizes these principles through instruction--observation symbolic encoding and hyperdimensional parameter reasoning, as detailed in the following sections.

\subsection{Overview of the ScanHD Framework}

\begin{figure*}[]
  \centering
  \includegraphics[width=\linewidth]{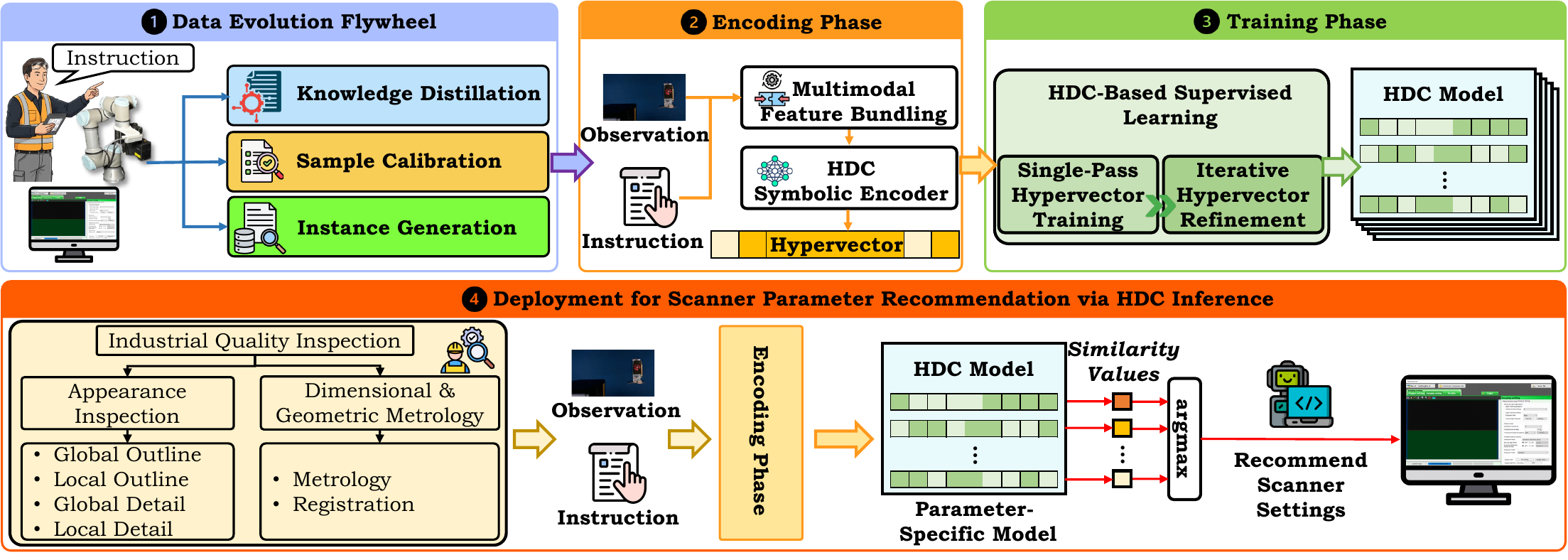}
  \caption{Overview of the proposed ScanHD framework.
ScanHD consists of four stages.
(1) Data Evolution Flywheel constructs high-quality instruction--observation--parameter training instances (with associated canonical intent labels) through knowledge distillation, sample calibration, and instance generation.
(2) Encoding Phase maps a visual observation and a natural-language instruction into a unified symbolic hypervector via multimodal feature bundling and hyperdimensional encoding.
(3) Training Phase learns parameter-specific associative memories using HDC-based supervised learning, combining single-pass hypervector training with optional iterative hypervector refinement to obtain compact and robust HDC models.
(4) Deployment Phase performs similarity-based HDC inference, where the encoded query hypervector is compared against learned class hypervectors to recommend scanner settings via efficient $\arg\max$ operations.}
  \label{fig:scanhd_model}
\end{figure*}

Figure~\ref{fig:scanhd_model} presents an overview of the proposed ScanHD framework.
ScanHD is designed as a hyperdimensional, instruction-conditioned sensing system that recommends scanner parameter configurations by explicitly separating data preparation, symbolic encoding, parameter learning, and deployment-time inference.

At a high level, ScanHD consists of four interconnected components, corresponding to the numbered stages in Fig.~\ref{fig:scanhd_model}.
First, a data evolution flywheel (\ding{182}) is used to construct a high-quality training corpus of instruction--observation--parameter instances $(t,o,\vartheta)$, together with their associated canonical intent labels $\mathbf{s}$ (Sec.~\ref{sec:data_flywheel}).
This process integrates knowledge distillation, sample calibration, and instance generation to ensure consistency between inspection intent, visual appearance, and scanner settings.
Details of the data evolution flywheel are provided in Sec.~\ref{sec:data_flywheel}.

Second, ScanHD performs instruction--observation symbolic encoding (\ding{183}), where a visual observation and a natural-language inspection instruction are jointly transformed into a unified hyperdimensional representation.
This encoding phase serves as the interface between multimodal inputs and downstream hyperdimensional reasoning, producing a task-aware symbolic state that captures both inspection intent and observable scene properties.

Third, ScanHD conducts hyperdimensional parameter learning (\ding{184}) using HDC-based supervised learning.
Each scanning parameter is modeled with a dedicated associative memory, allowing parameter-wise reasoning while sharing a common symbolic representation.
Through hypervector label assembly and multi-epoch training, ScanHD learns stable class prototypes that reflect consistent sensing behaviors across diverse observations and instructions.

Finally, during deployment, ScanHD performs parameter recommendation via HDC inference (\ding{185}).
Given a new observation and instruction, the encoded hypervector is compared against the learned parameter-specific memories using similarity-based retrieval.
The resulting similarity scores are used to infer discrete scanner settings through lightweight $\arg\max$ operations, enabling real-time and interpretable sensor configuration for industrial quality inspection tasks.

In the following subsections, we describe each component of the ScanHD framework in detail, following the same order as illustrated in Fig.~\ref{fig:scanhd_model}.

\subsubsection{Encoding Phase: Instruction--Observation Symbolic Representation}
\label{sec:encoding_phase}

ScanHD follows a representation-first design in which multimodal inputs are first mapped into a shared embedding space and subsequently transformed into a symbolic hyperdimensional representation suitable for parameter reasoning.
The encoding phase (\ding{183} in Fig.~\ref{fig:scanhd_model}) consists of three steps: generating multimodal embeddings using a pretrained vision language model, binding instruction and observation embeddings to obtain a task-conditioned representation, and performing symbolic encoding via HDC projection.

\noindent \textit{Generating Multimodal Embeddings.}
Given a visual observation $o$ and a natural-language inspection instruction $t$, we first employ a pretrained vision language model, specifically CLIP, to extract their continuous embeddings.
The observation is mapped to a visual embedding
\begin{equation}
\mathbf{e}_o = \mathrm{CLIP}_{\text{image}}(o)
\end{equation}
and the instruction is mapped to a textual embedding
\begin{equation}
\mathbf{e}_t = \mathrm{CLIP}_{\text{text}}(t)
\end{equation}
where both $\mathbf{e}_o$ and $\mathbf{e}_t$ lie in a shared semantic space.

\noindent \textit{Instruction--Observation Feature Binding.}
To align perception with inspection intent, ScanHD binds the observation embedding with the instruction embedding to form an instruction-conditioned representation: $\mathbf{z} = \mathcal{B}(\mathbf{e}_o, \mathbf{e}_t)$, where $\mathcal{B}(\cdot)$ represents a feature binding operator that integrates visual cues with task semantics.
Intuitively, this step emphasizes aspects of the observation that are relevant to the specified inspection objective, while suppressing task-irrelevant appearance variations.
The resulting representation $\mathbf{z}$ encodes how the scene should be sensed under the given instruction, rather than what is present in the scene.

\noindent \textit{Hyperdimensional Symbolic Encoding.}
The instruction-conditioned representation $\mathbf{z}$ is subsequently transformed into a symbolic hypervector using a HDC encoder:
\begin{equation}
\mathbf{h} = \mathcal{E}(\mathbf{z}) \in \{-1, +1\}^{D_h}
\end{equation}
where $\mathcal{E}(\cdot)$ denotes the HDC symbolic encoding function.
This projection converts continuous multimodal features into a distributed symbolic representation that preserves similarity relationships through quasi-orthogonality in high-dimensional space.

Although the instruction $t$ is provided as a single natural-language sentence, the resulting hypervector $\mathbf{h}$ is symbolic in nature.
Multiple semantic factors jointly contribute to $\mathbf{h}$ through HDC superposition, enabling compositional and interpretable reasoning in subsequent stages.
Importantly, the encoding phase is agnostic to individual scanning parameters; it produces a shared symbolic representation that serves as the input to parameter-specific HDC learning modules described next.

\subsubsection{Training Phase: HDC-Based Parameter Learning}
\label{sec:training_phase}

The goal of the training phase (\ding{184} in Fig.~\ref{fig:scanhd_model}) is to construct compact associative memories that map instruction-conditioned hypervectors to discrete scanning parameter choices.
A key challenge in ScanHD is that many training instances share highly similar sensing patterns (e.g., common appearances or recurring instruction intents), while a smaller portion of instances correspond to long-tail cases that nevertheless matter for robust industrial deployment.
Naively bundling all encoded hypervectors can over-emphasize frequent patterns and lead to saturation of class hypervectors, reducing the model's ability to represent rare but informative variations.
Motivated by prior work on adaptive online hyperdimensional learning, such as OnlineHD~\cite{hernandez2021onlinehd}, ScanHD adopts an adaptive learning procedure that (i) learns stable class prototypes in a single pass and (ii) optionally refines class boundaries through lightweight error-driven iterations.

\noindent \textit{Parameter-Specific Associative Memories.}
Let $P$ denote the number of controllable scanning parameters. For each parameter index $k \in \{1,2,\dots,P\}$, let $\mathcal{Y}_k$ denote its discrete value set.
ScanHD trains an independent associative memory $\mathcal{M}_k$ for each parameter, represented by a set of class hypervectors $\{\mathbf{C}_{k,v}\}_{v\in\mathcal{Y}_k}$.
All parameters share the same encoded instruction--observation hypervector $\mathbf{h}$ (Sec.~\ref{sec:encoding_phase}), but maintain separate class memories to avoid entangling physically distinct sensing decisions.

\noindent \textit{Single-Pass Training.}
ScanHD first builds a strong initial model by processing the training data once.
For a training instance encoded as $\mathbf{h}$ with ground-truth parameter value $\vartheta_k \in \mathcal{Y}_k$, we compute cosine similarity to each class hypervector:
\begin{equation}
\hat{\vartheta}_k = \arg\max_{v \in \mathcal{Y}_k} \; \delta(\mathbf{h}, \mathbf{C}_{k,v})
\end{equation}
where $\delta(\cdot,\cdot)$ denotes cosine similarity.
Instead of always adding $\mathbf{h}$ to the class hypervector, ScanHD updates the model proportionally to the novelty of the sample with respect to the current class representation.
Concretely, we update the correct class hypervector using a similarity-modulated rule:
\begin{equation}
\mathbf{C}_{k,\vartheta_k} \leftarrow \mathbf{C}_{k,\vartheta_k}
+ \eta \left(1 - \delta(\mathbf{h}, \mathbf{C}_{k,\vartheta_k})\right)\mathbf{h}
\end{equation}
where $\eta$ is a learning rate.
When $\mathbf{h}$ is already well represented by $\mathbf{C}_{k,\vartheta_k}$ (high similarity), the update magnitude is small, preventing saturation and over-counting of common patterns.
When $\mathbf{h}$ introduces new structure (low similarity), the update becomes larger, ensuring that informative variations are incorporated even if they appear infrequently.
This yields a well-trained initial associative memory after a single pass.

\noindent \textit{Iterative Hypervector Refinement.}
Starting from the adaptive single-pass model, ScanHD optionally performs a small number of refinement epochs to sharpen decision boundaries.
During refinement, each training hypervector is classified using the current associative memory; updates are applied only when the sample is misclassified.
If a sample $\mathbf{h}$ with true value $\vartheta_k$ is predicted as $\hat{\vartheta}_k \neq \vartheta_k$, ScanHD reinforces the correct class and penalizes the mis-predicted class using similarity-aware updates:
\begin{equation}
\mathbf{C}_{k,\vartheta_k} \leftarrow \mathbf{C}_{k,\vartheta_k}
+ \eta \left(1 - \delta(\mathbf{h}, \mathbf{C}_{k,\vartheta_k})\right)\mathbf{h}
\end{equation}
\begin{equation}
\mathbf{C}_{k,\hat{\vartheta}_k} \leftarrow \mathbf{C}_{k,\hat{\vartheta}_k}
- \eta \left(1 - \delta(\mathbf{h}, \mathbf{C}_{k,\hat{\vartheta}_k})\right)\mathbf{h}
\end{equation}

This refinement stage focuses computation on hard cases and long-tail confusions, and it converges quickly because it starts from a strong adaptive initialization and only corrects residual misclassifications.

\subsubsection{Deployment Phase: Parameter Recommendation via HDC Inference}
\label{sec:deployment_phase}

At deployment time, ScanHD performs efficient scanner parameter recommendation via similarity-based hyperdimensional inference.
The deployment phase (\ding{185} in Fig.~\ref{fig:scanhd_model}) takes as input a new visual observation and an inspection instruction, and outputs a discrete configuration for each scanning parameter using the learned associative memories.

\noindent \textit{Instruction--Observation Encoding.}
Given a query observation $o$ and instruction $t$, ScanHD first applies the same encoding pipeline described in Sec.~\ref{sec:encoding_phase}.
Specifically, the observation and instruction are embedded using the pretrained vision language model, bound into an instruction-conditioned representation, and projected into a hypervector.

\noindent \textit{Similarity-Based Parameter Inference.}
For each scanning parameter index $k \in \{1,2,\dots,P\}$, ScanHD retrieves the corresponding associative memory $\mathcal{M}_k$, which consists of class hypervectors $\{\mathbf{C}_{k,v}\}_{v \in \mathcal{Y}_k}$.
The similarity between the query hypervector and class hypervector is computed as
\begin{equation}
s_{k,v} = \delta(\mathbf{h}, \mathbf{C}_{k,v})
\end{equation}

The recommended parameter value is obtained via a maximum-similarity decision:
\begin{equation}
\hat{\vartheta}_k = \arg\max_{v \in \mathcal{Y}_k} \; s_{k,v}
\end{equation}

This inference process is repeated independently for each scanning parameter, yielding the full recommended configuration $\hat{\vartheta} = (\hat{\vartheta}_1,\hat{\vartheta}_2,\dots,\hat{\vartheta}_P)$.

\noindent \textit{Efficiency and Interpretability.}
Because inference relies only on vector similarity and $\arg\max$ operations, ScanHD incurs minimal computational overhead and is well suited for real-time deployment.
Moreover, the similarity scores $s_{k,v}$ provide an interpretable measure of confidence for each parameter choice, enabling inspection of how the instruction-conditioned sensing state aligns with different parameter settings.
This lightweight and transparent inference mechanism allows ScanHD to adapt scanner configurations on the fly while maintaining robustness across objects, tasks, and appearance conditions.


\section{Experimental Results} \label{ExperimentalResult}

\subsection{Evaluation Metrics}
We evaluate scanning parameter recommendation using three metrics: Exact Accuracy, Win@1 Accuracy, and F1 score. Each metric is defined at the level of individual predictions and reflects a different notion of correctness under discretized scanning parameters.

Let $y$ denote the ground-truth discrete value of a scanning parameter and $\hat{y}$ denote the predicted value.

Exact Accuracy evaluates whether the predicted parameter exactly matches the ground truth:
\begin{equation}
\mathrm{Exact}(y, \hat{y}) =
\begin{cases}
1, & \text{if } \hat{y} = y \\
0, & \text{otherwise}
\end{cases}
\end{equation}
Exact Accuracy is computed as the average of $\mathrm{Exact}(y, \hat{y})$ over all evaluation samples.

Win@1 Accuracy evaluates whether the predicted parameter falls within one adjacent discrete level of the ground truth. Let $\mathrm{ord}(\cdot)$ denote the ordinal index of a parameter value in its predefined ordered set:
\begin{equation}
\mathrm{Win@1}(y, \hat{y}) =
\begin{cases}
1, & \text{if } \left| \mathrm{ord}(\hat{y}) - \mathrm{ord}(y) \right| \leq 1 \\
0, & \text{otherwise}
\end{cases}
\end{equation}
Win@1 Accuracy captures near-correct predictions that may still yield acceptable scanning quality in practice. However, for Measurement Range (X), adjacent discrete levels correspond to substantially different fields of view, and being off by one can lead to severe clipping or loss of target geometry; therefore, Win@1 Accuracy is not reported for Measurement Range (X).

F1 Score evaluates prediction quality by jointly considering false positives and false negatives across discrete parameter values. For each parameter, the final F1 score is computed by aggregating binary prediction outcomes across the evaluation set and averaging across classes to avoid bias toward frequent values.

\subsection{Main Results}

\begin{table*}[t]
\centering
\caption{\textbf{Main Benchmark Results on the Instruct-Obs2Param Dataset.}
Each model predicts 5 Keyence LJ-X8200 scanning parameters. 
For each parameter we report Exact Accuracy, Win-1 Accuracy, and F1.}
\label{tab:main_results}
\resizebox{\textwidth}{!}{
\begin{tabular}{l|ccc|ccc|ccc|ccc|ccc|ccc|ccc}
\toprule
& \multicolumn{3}{c|}{\textbf{Sample Freq}} 
& \multicolumn{3}{c|}{\textbf{Measurement Range X}} 
& \multicolumn{3}{c|}{\textbf{Exposure Time}} 
& \multicolumn{3}{c|}{\textbf{CMOS Dynamic Range}} 
& \multicolumn{3}{c|}{\textbf{Light Intensity Range}} 
& \multicolumn{3}{c}{\textbf{Average}} \\
\textbf{Method} 
& Exact & Win@1 & F1
& Exact & Win@1 & F1
& Exact & Win@1 & F1
& Exact & Win@1 & F1
& Exact & Win@1 & F1
& Exact & Win@1 & F1 \\ 
\midrule
\multicolumn{19}{l}{\textbf{A. Rule-based Baselines}} \\
Rule-based Heuristic 
& 39.9$\pm$1.8 & 59.1$\pm$2.6 & 23.3$\pm$1.7 
& 56.5$\pm$4.7 & - & 24.0$\pm$1.3 
& 47.7$\pm$2.1 & 78.4$\pm$2.7 & 21.9$\pm$1.2 
& 53.1$\pm$3.0 & 89.2$\pm$2.4 & 25.3$\pm$0.9 
& 54.7$\pm$4.7 & 100.0$\pm$0.0 & 23.8$\pm$1.7 
& 50.4$\pm$3.3 & 81.7$\pm$1.5 & 23.7$\pm$1.4\\
\midrule

\multicolumn{19}{l}{\textbf{B. Instruction-Only Methods}} \\
Logistic Regression
& \textbf{97.9$\pm$1.4} & \textbf{98.6$\pm$0.9} & \textbf{98.0$\pm$1.4} 
& 93.1$\pm$1.7 & - & 91.8$\pm$1.7 
& 75.3$\pm$2.3 & 92.7$\pm$1.3 & 74.1$\pm$2.3 
& 83.3$\pm$2.1 & 96.5$\pm$0.5 & 81.5$\pm$2.5 
& 90.2$\pm$0.7 & 95.5$\pm$1.1 & 88.3$\pm$1.2 
& 88.0$\pm$1.6 & 95.8$\pm$0.5 & 86.7$\pm$1.8 \\

KNN
& 92.6$\pm$2.3 & 95.7$\pm$2.4 & 92.6$\pm$2.0 
& 83.4$\pm$1.2 & - & 82.1$\pm$2.1 
& 74.8$\pm$1.5 & 91.9$\pm$1.6 & 73.3$\pm$1.3 
& 80.9$\pm$2.6 & 95.5$\pm$1.3 & 77.7$\pm$3.8 
& 79.3$\pm$1.4 & 93.5$\pm$1.2 & 75.2$\pm$0.8 
& 82.2$\pm$1.8 & 94.1$\pm$1.6 & 80.2$\pm$2.0 \\


\midrule

\multicolumn{19}{l}{\textbf{C. Observation-Only Methods}} \\
Resnet
& 42.2$\pm$3.6 & 62.6$\pm$3.6 & 28.5$\pm$3.4 
& 71.2$\pm$2.1 & - & 70.5$\pm$2.2 
& 74.8$\pm$2.4 & \textbf{100.0$\pm$0.0} & 73.7$\pm$3.2 
& 73.7$\pm$3.1 & \textbf{100.0$\pm$0.0} & 74.8$\pm$4.8 
& \textbf{100.0$\pm$0.0} & \textbf{100.0$\pm$0.0} & \textbf{100.0$\pm$0.0} 
& 72.4$\pm$2.2 & 90.7$\pm$1.8 & 69.5$\pm$2.7 \\

ViT
& 44.2$\pm$3.1 & 65.0$\pm$3.7 & 31.3$\pm$4.5 
& 71.4$\pm$1.7 & - & 72.2$\pm$2.8 
& 72.2$\pm$2.9 & \textbf{100.0$\pm$0.0} & 72.1$\pm$2.4 
& 74.1$\pm$2.7 & \textbf{100.0$\pm$0.0} & 77.0$\pm$2.1 
& \textbf{100.0$\pm$0.0} & \textbf{100.0$\pm$0.0} & \textbf{100.0$\pm$0.0} 
& 72.4$\pm$2.1 & 91.3$\pm$1.8 & 70.5$\pm$2.4 \\

\midrule

\multicolumn{19}{l}{\textbf{D. Fusion Methods}} \\
Early Fusion
& 85.1$\pm$3.8 & 90.9$\pm$0.9 & 84.2$\pm$6.2 
& 91.2$\pm$4.3 & - & 91.6$\pm$4.5 
& 84.5$\pm$2.8 & 100.0$\pm$0.0 & 85.2$\pm$2.8 
& 86.2$\pm$0.8 & 99.8$\pm$0.5 & \textbf{85.6$\pm$1.3} 
& 98.8$\pm$0.6 & 99.4$\pm$0.4 & 98.6$\pm$0.7 
& 89.2$\pm$2.5 & 97.5$\pm$0.6 & 89.0$\pm$3.1\\

Late Fusion
& 76.0$\pm$1.3 & 89.2$\pm$1.2 & 70.4$\pm$1.6 
& 88.0$\pm$3.4 & - & 86.5$\pm$6.6 
& 76.2$\pm$2.8 & 99.9$\pm$0.2 & 75.6$\pm$3.0 
& 75.1$\pm$2.9 & 94.2$\pm$2.4 & 68.0$\pm$10.3 
& 98.7$\pm$0.6 & 99.1$\pm$0.6 & 98.2$\pm$0.9 
& 82.8$\pm$2.2 & 95.6$\pm$0.8 & 79.7$\pm$4.5\\

\midrule

\multicolumn{19}{l}{\textbf{D. Multimodal Large Language Models}} \\
Qwen3-VL-4B-Instruct
& 60.0$\pm$4.7 & 92.8$\pm$1.3 & 55.4$\pm$5.0 
& 59.3$\pm$1.6 & - & 48.7$\pm$2.3 
& 24.0$\pm$2.8 & 89.2$\pm$3.2 & 18.9$\pm$2.8 
& 65.6$\pm$2.4 & 94.3$\pm$1.1 & 53.5$\pm$3.4 
& 40.2$\pm$5.2 & 79.7$\pm$3.8 & 36.6$\pm$4.9 
& 49.8$\pm$3.3 & 89.0$\pm$1.8 & 42.6$\pm$3.7\\

Qwen3-VL-4B-Thinking
& 62.0$\pm$2.7 & 84.3$\pm$2.2 & 59.5$\pm$2.3 
& 60.0$\pm$2.5 & - & 47.1$\pm$2.3 
& 36.7$\pm$1.2 & 80.2$\pm$1.0 & 31.3$\pm$1.4 
& 52.9$\pm$1.8 & 84.3$\pm$1.0 & 43.2$\pm$2.0 
& 27.8$\pm$1.2 & 78.4$\pm$2.0 & 27.7$\pm$1.3 
& 47.9$\pm$1.9 & 81.8$\pm$1.1 & 41.8$\pm$1.9 \\

Qwen3-VL-8B-Instruct
& 64.3$\pm$1.1 & 89.5$\pm$1.3 & 48.9$\pm$1.1 
& 63.3$\pm$1.9 & - & 46.0$\pm$2.0 
& 34.2$\pm$1.2 & 69.3$\pm$0.9 & 27.2$\pm$1.9 
& 53.1$\pm$2.8 & 87.1$\pm$2.4 & 42.9$\pm$2.0 
& 34.9$\pm$2.6 & 78.5$\pm$1.6 & 30.1$\pm$2.3 
& 49.9$\pm$1.9 & 81.1$\pm$1.1 & 39.0$\pm$1.9  \\

Qwen3-VL-8B-Thinking
& 62.6$\pm$3.3 & 86.9$\pm$1.5 & 57.5$\pm$3.7 
& 61.4$\pm$1.4 & - & 45.8$\pm$1.0 
& 33.3$\pm$1.6 & 84.1$\pm$3.3 & 33.2$\pm$1.7 
& 47.6$\pm$2.0 & 82.0$\pm$2.1 & 40.6$\pm$1.9 
& 43.3$\pm$2.0 & 82.7$\pm$1.5 & 33.8$\pm$2.0 
& 49.6$\pm$2.1 & 83.9$\pm$1.2 & 42.2$\pm$2.1\\

\midrule

\multicolumn{19}{l}{\textbf{Our Model}} \\

\textbf{ScanHD}
& 90.6$\pm$2.6 & 94.3$\pm$1.4 & 90.5$\pm$2.7 
& \textbf{96.5$\pm$1.9} & - & \textbf{97.1$\pm$1.5} 
& \textbf{90.6$\pm$1.9} & 99.7$\pm$0.5 & \textbf{90.3$\pm$1.9} 
& \textbf{86.4$\pm$0.7} & 99.9$\pm$0.2 & 81.5$\pm$1.7 
& 99.3$\pm$0.7 & 100.0$\pm$0.0 & 99.4$\pm$0.5 
& \textbf{92.7$\pm$1.6} & \textbf{98.5$\pm$0.4} & \textbf{91.8$\pm$1.7}\\
\bottomrule
\end{tabular}
}
\end{table*}

To provide a comprehensive and fair comparison, we evaluate ScanHD against a diverse set of baselines that span different modeling assumptions and sources of information. Specifically, the evaluated methods include rule-based instruction lookup strategies, instruction-only learning models, observation-only visual models, multimodal fusion approaches, and general-purpose multimodal large language models. This coverage allows us to systematically analyze how sensing parameter recommendation is affected by inspection intent, visual appearance, and inductive bias.

The rule-based baseline is implemented as an instruction-driven lookup strategy rather than physics-specific expert heuristics. Each instruction is normalized through lowercasing and light punctuation processing. A lookup table is constructed on the training split by grouping samples with identical normalized instructions and assigning each scanning parameter the modal value observed for that instruction. At test time, if an instruction has been seen during training, the corresponding modal configuration is retrieved; otherwise, a global per-parameter mode computed from the training set is used as a fallback. This baseline represents a strong instruction-level memorization strategy under row-random splits.

\begin{figure}[]
  \centering
  \includegraphics[width=\linewidth]{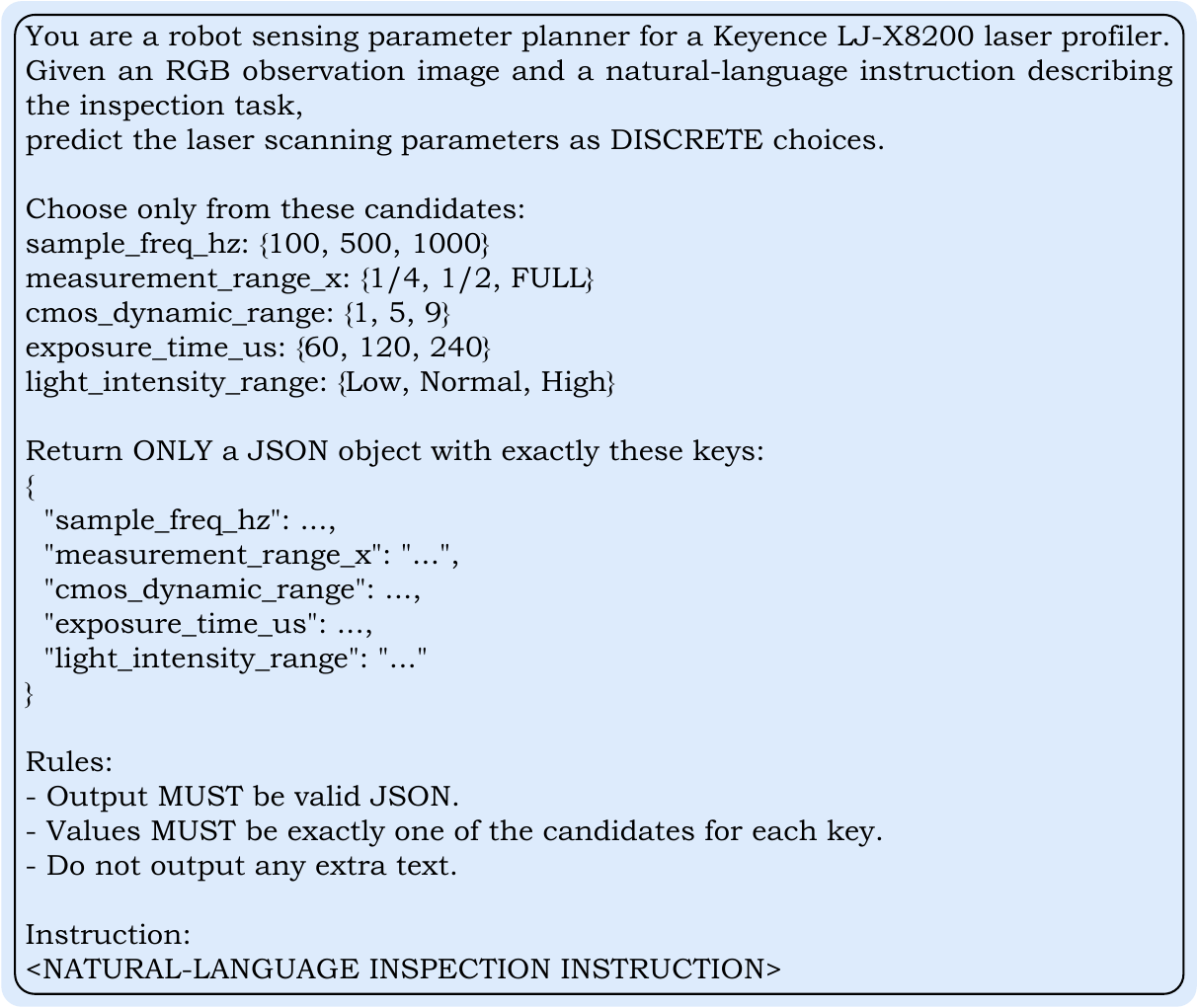}
  \caption{Prompt template used to evaluate the multimodal large language models for laser scanning parameter prediction.}
  \label{qwen_prompt}
\end{figure}

Instruction-only methods take the natural-language instruction as the sole input and ignore visual observations. We evaluate Logistic Regression and a k-nearest neighbor (KNN) classifier operating on CLIP text embeddings to quantify how much scanning parameter information can be inferred purely from inspection intent. Observation-only methods rely exclusively on visual input and disregard the instruction; ResNet and ViT models trained on RGB observations are used to assess the contribution of appearance cues alone. Fusion-based methods combine instruction and observation embeddings using two common strategies: Early Fusion concatenates multimodal features before prediction, while Late Fusion performs independent predictions from each modality and fuses decisions at the output level. In addition, we evaluate multimodal large language models using Qwen3-VL \cite{Qwen3-VL} variants with different model scales (4B and 8B) and reasoning modes (Instruct and Thinking). These models are evaluated in a zero-shot setting and are prompted to directly output discrete scanning parameters from the image and instruction, without any task-specific fine-tuning or structural constraints. The prompt used for multimodal large language model evaluation is illustrated in Figure~\ref{qwen_prompt}.

\begin{figure}[]
  \centering
  \includegraphics[width=\linewidth]{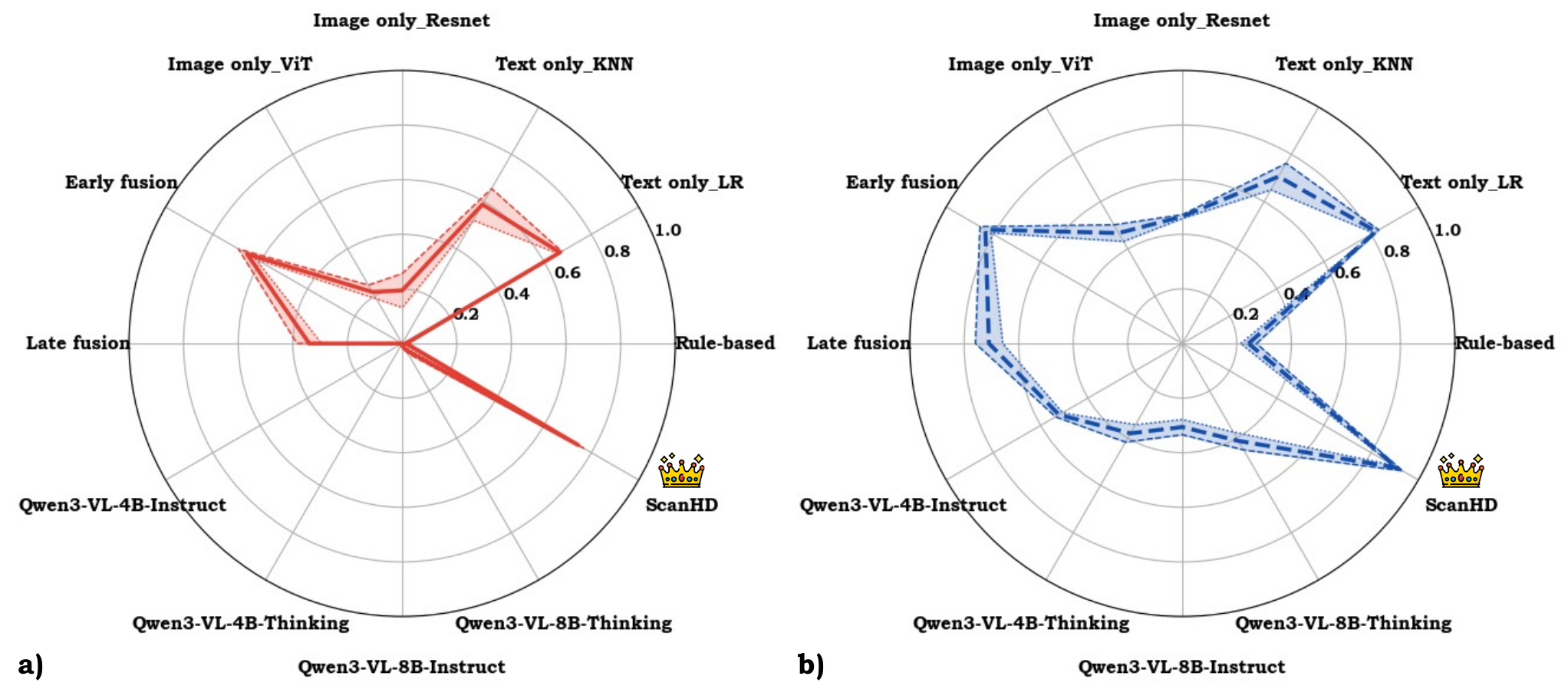}
  \caption{System-level evaluation of all-parameter correctness across different methods. A prediction is considered correct only if all five scanning parameters are inferred correctly for a given instruction--observation pair. Radar plots report (a) Exact Accuracy and (b) Win@1 Accuracy under this all-parameter criterion.}
  \label{exp: main result}
\end{figure}

Table~\ref{tab:main_results} reports per-parameter results across all methods. Beyond aggregate performance gains, the results reveal systematic failure modes of unimodal and naive fusion baselines, clarifying where ScanHD provides decisive advantages.
A first observation is the complementary yet insufficient nature of unimodal reasoning. Instruction-only methods perform competitively on parameters primarily driven by inspection intent. For example, Logistic Regression and KNN achieve over $92\%$ Exact Accuracy on sampling frequency and over $83\%$ on measurement range, reflecting strong instruction-level regularities. However, their performance degrades on appearance-sensitive parameters. On exposure time, Exact Accuracy drops to approximately $75\%$, indicating limited sensitivity to visual cues such as surface reflectivity and illumination.
Observation-only methods exhibit the opposite failure pattern. While achieving near-perfect accuracy on light intensity range, their Exact Accuracy on sampling frequency remains below $45\%$, reflecting difficulty in inferring intent-dependent decisions from appearance alone. This asymmetric behavior highlights that neither modality in isolation can resolve the full decision space, as inspection intent and visual appearance govern different subsets of scanning parameters.
Fusion-based approaches partially mitigate these issues but remain constrained by their lack of parameter-specific structure. Early fusion achieves an average Exact Accuracy of $89.2\%$, and late fusion reaches $82.8\%$, both consistently trailing ScanHD across all five parameters. These results suggest that naive feature-level or decision-level fusion is insufficient to capture the heterogeneous dependencies between instruction semantics, observation cues, and parameter-specific decision logic.
In contrast, ScanHD maintains consistently high performance across both intent-dominated and appearance-sensitive parameters, achieving an average Exact Accuracy of $92.7\%$, Win@1 Accuracy of $98.5\%$, and an F1 score of $91.8\%$. Its advantage is most pronounced on mixed-dependency parameters, such as exposure time and CMOS dynamic range, where both inspection intent and visual appearance jointly constrain feasible parameter values.
The instruction-lookup rule-based baseline further illustrates this limitation. Although it attains perfect accuracy on sampling frequency ($100.0\%$ Exact Accuracy) due to strong instruction regularities, its performance degrades substantially on parameters requiring visual grounding, yielding an average Exact Accuracy of only $50.4\%$. Finally, multimodal large language models perform poorly across all evaluated parameters, with average Exact Accuracy below $50\%$, underscoring the difficulty of mapping free-form multimodal reasoning to precise, hardware-constrained scanning parameters without domain-specific inductive bias or structured memory.

While per-parameter performance provides useful insight, practical robotic inspection requires all scanning parameters to be correct simultaneously. Figure~\ref{exp: main result} therefore reports system-level results under the all-parameter correctness criterion, where a prediction is considered correct only if all five parameters match the ground truth. 
For Figure~\ref{exp: main result}(b), Measurement Range (X) is required to be exactly correct, rather than within a Win@1 tolerance, since even off-by-one errors can lead to catastrophic sensing failure.
Under this stricter evaluation, ScanHD demonstrates a clear and consistent advantage over all baselines. The performance gap widens substantially compared to per-parameter metrics, revealing that many competing methods suffer from error accumulation and inconsistent parameter combinations.

\subsection{Cross-split Generalization Study}
\begin{table*}[]
\centering
\caption{Cross-split generalization performance of \textbf{ScanHD} on Position, Rotation, Lightingsplits.
For each parameter, we report Exact Accuracy, Win@1 Accuracy, and F1 score (\%), together with averaged metrics across all parameters.}
\label{tab:scanhd_cross_split}
\resizebox{\linewidth}{!}{
\setlength{\tabcolsep}{3pt}
\renewcommand{\arraystretch}{1.15}
\begin{tabular}{l|
ccc ccc ccc ccc ccc|ccc}
\toprule
\multirow{2}{*}{\textbf{Split}} 
& \multicolumn{3}{c}{\textbf{Sample Freq}} 
& \multicolumn{3}{c}{\textbf{Measurement Range X}} 
& \multicolumn{3}{c}{\textbf{Exposure Time}} 
& \multicolumn{3}{c}{\textbf{CMOS Dynamic Range}} 
& \multicolumn{3}{c}{\textbf{Light Intensity Range}} 
& \multicolumn{3}{c}{\textbf{Average}} \\
\cmidrule(lr){2-16}\cmidrule(lr){17-19}
& Exact & Win@1 & F1
& Exact & Win@1 & F1
& Exact & Win@1 & F1
& Exact & Win@1 & F1
& Exact & Win@1 & F1
& Exact & Win@1 & F1 \\
\midrule
\textbf{Position} 
& 97.5$\pm$0.6 & 98.5$\pm$0.5 & 97.4$\pm$0.6 
& 96.2$\pm$1.7 & - & 92.8$\pm$4.3
& 93.6$\pm$2.8 & 99.6$\pm$0.4 & 93.8$\pm$2.7 
& 89.1$\pm$3.9 & 98.8$\pm$0.8 & 83.4$\pm$7.8 
& 96.8$\pm$2.4 & 97.6$\pm$1.7 & 94.9$\pm$3.6 
& 94.6$\pm$2.3 & 98.6$\pm$0.5 & 92.5$\pm$3.8 \\
\textbf{Rotation} 
& 98.0$\pm$0.4 & 98.8$\pm$0.3 & 98.0$\pm$0.4 
& 97.7$\pm$0.6 & - & 96.4$\pm$1.1 
& 94.5$\pm$0.7 & 99.7$\pm$0.1 & 94.5$\pm$0.7 
& 91.8$\pm$1.0 & 99.2$\pm$0.3 & 89.4$\pm$2.3 
& 98.1$\pm$0.3 & 98.6$\pm$0.4 & 97.2$\pm$0.6 
& 96.0$\pm$0.6 & 99.1$\pm$0.1 & 95.1$\pm$1.0 \\
\textbf{Lighting} 
& 95.5$\pm$1.5 & 97.4$\pm$1.2 & 95.5$\pm$1.5 
& 92.6$\pm$2.9 & - & 84.5$\pm$9.2 
& 90.1$\pm$2.8 & 99.3$\pm$0.5 & 90.2$\pm$2.8 
& 85.4$\pm$5.7 & 98.5$\pm$1.0 & 80.0$\pm$9.8 
& 94.4$\pm$3.6 & 95.4$\pm$3.4 & 89.5$\pm$8.5 
& 91.6$\pm$3.3 & 97.7$\pm$0.9 & 87.9$\pm$6.3 \\
\bottomrule
\end{tabular}
}
\end{table*}

\begin{figure*}[]
  \centering
  \includegraphics[width=\linewidth]{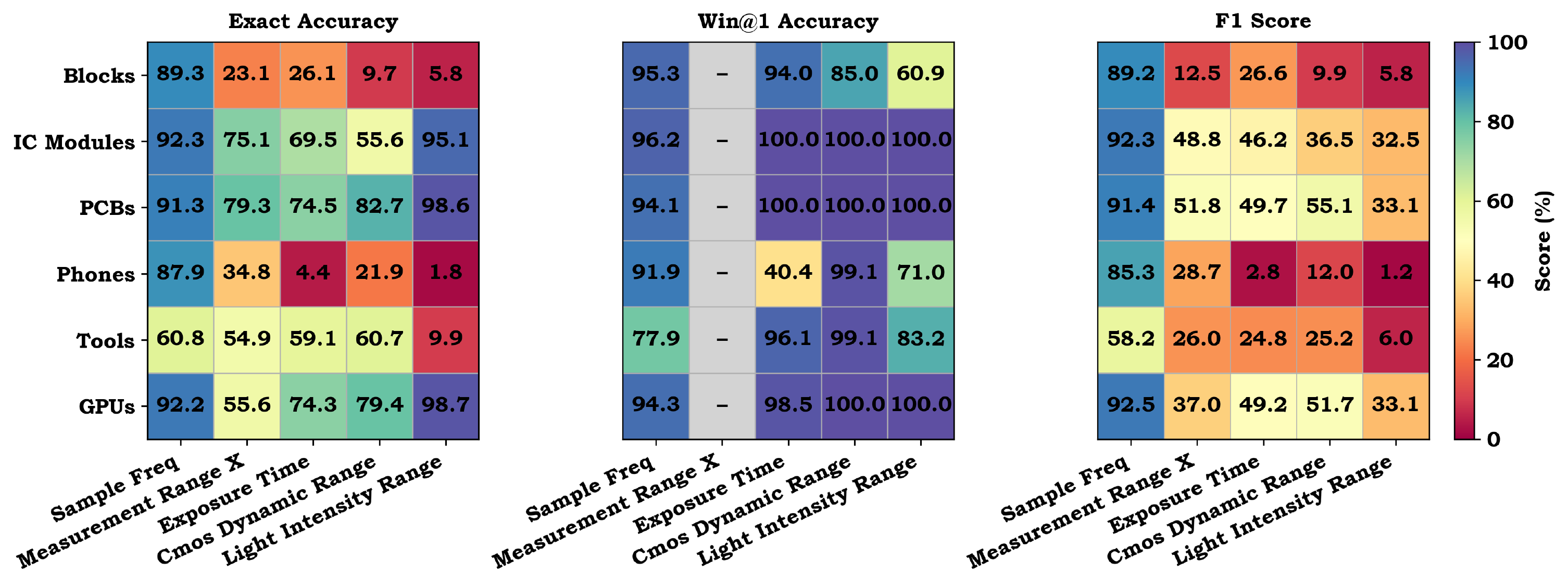}
  \caption{Category-wise analysis of ScanHD across scanning parameters. Heatmaps show Exact Accuracy, Win@1 Accuracy, and F1 score, respectively. }
  \label{exp_object_split}
\end{figure*}

We further evaluate the generalization ability of ScanHD under cross-split settings that isolate different sources of distribution shift, including position, rotation, and lighting. In each setting, the model is trained on a subset of conditions and evaluated on held-out splits, enabling a direct assessment of robustness to viewpoint and illumination variations that commonly arise in real-world robotic inspection.

Table~\ref{tab:scanhd_cross_split} summarizes the cross-split performance of ScanHD across all scanning parameters. Overall, ScanHD maintains consistently strong performance across all three splits. Among them, the rotation split yields the highest average performance, with an average Exact Accuracy of $96.0\%$, Win@1 Accuracy of $99.1\%$, and F1 score of $95.1\%$. Performance under position and lighting splits remains comparably high, with average Exact Accuracy above $91\%$ and Win@1 Accuracy exceeding $97\%$, indicating that ScanHD generalizes well across changes in sensor pose and illumination conditions.

Across individual parameters, ScanHD demonstrates robust behavior under all splits. Sampling frequency and measurement range remain highly stable, with Exact Accuracy consistently above $90\%$ across position, rotation, and lighting splits. Appearance-sensitive parameters, including exposure time and light intensity range, exhibit slightly larger performance variation, particularly under lighting shifts. Nevertheless, Win@1 Accuracy for these parameters remains above $94\%$ in all cases, suggesting that most errors correspond to near-correct predictions within adjacent discrete levels.

Figure~\ref{exp_object_split} provides a category-wise analysis of ScanHD across different object classes. The heatmaps reveal distinct parameter-dependent generalization patterns. For object categories with complex geometry and heterogeneous materials, such as IC modules, PCBs, and GPUs, ScanHD achieves strong Exact and Win@1 Accuracy across all parameters, indicating effective transfer of instruction-conditioned representations. In contrast, categories such as phones and tools show lower Exact Accuracy on appearance-sensitive parameters, especially exposure time and light intensity range, reflecting increased sensitivity to surface reflectivity and lighting conditions. Importantly, these drops are largely mitigated under the Win@1 criterion, highlighting the practical robustness of ScanHD in producing usable scanning configurations despite challenging visual conditions.

Taken together, the cross-split results demonstrate that ScanHD does not rely on memorization of specific viewpoints or lighting setups. Instead, it learns a structured instruction-conditioned sensing representation that generalizes across spatial and illumination variations while maintaining high system-level consistency.

\subsection{Data Efficiency under Limited Supervision}

\begin{figure*}[t]
  \centering
  \includegraphics[width=\linewidth]{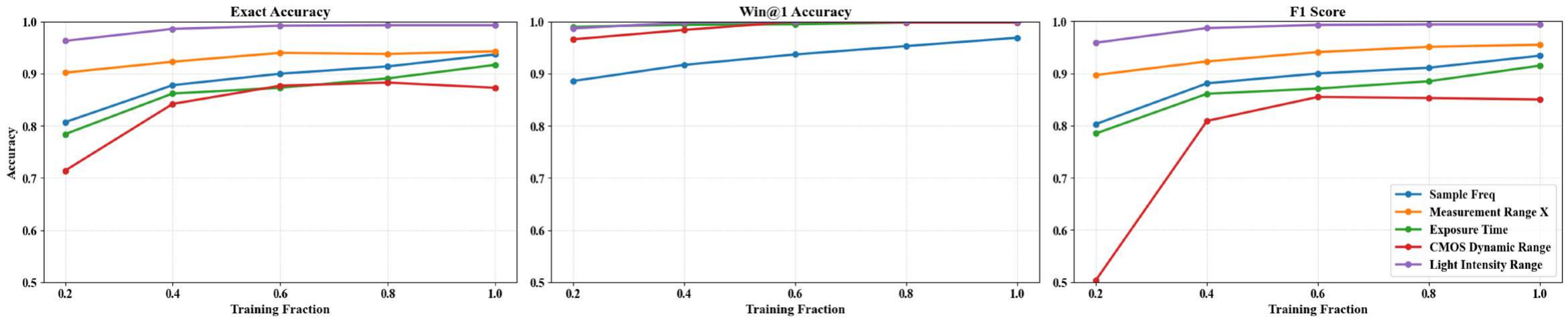}
  \caption{Data efficiency of ScanHD under limited supervision. Performance is evaluated by varying the fraction of training data from $20\%$ to $100\%$. Curves report Exact Accuracy, Win@1 Accuracy, and F1 score for each scanning parameter. }
  \label{exp: training ratio}
\end{figure*}

We evaluate the data efficiency of ScanHD by progressively reducing the amount of training data and analyzing model performance under limited supervision. The training fraction is varied from $20\%$ to $100\%$, while evaluation is performed on a fixed test set. Figure~\ref{exp: training ratio} reports Exact Accuracy, Win@1 Accuracy, and F1 score for each scanning parameter.

Across all parameters, ScanHD maintains strong performance even with limited training data. With only $20\%$ of the training samples, Win@1 Accuracy already exceeds $90\%$ for all parameters. As the training fraction increases, performance improves in a smooth and monotonic manner and reaches saturation without abrupt changes. This behavior indicates stable learning dynamics as additional supervision becomes available.

Parameters that are primarily driven by inspection intent, such as sampling frequency and measurement range, converge rapidly under limited supervision. Their Exact Accuracy exceeds $85\%$ when $40\%$ of the training data is used and approaches saturation beyond $60\%$. This trend shows that ScanHD can exploit instruction semantics effectively even when labeled data are scarce. In contrast, appearance-sensitive parameters, including exposure time and CMOS dynamic range, exhibit larger performance gains as the training set expands. These parameters require broader visual coverage, which explains their stronger dependence on increased data availability.

For all parameters, Win@1 Accuracy saturates earlier than Exact Accuracy and remains close to $100\%$ even in low-data regimes. This pattern implies that most errors under limited supervision correspond to near-correct parameter selections rather than severe misconfigurations. F1 score follows a similar trajectory and improves steadily as additional data are introduced, particularly for appearance-sensitive parameters where class imbalance is more pronounced.

Overall, these results demonstrate that ScanHD achieves strong data efficiency and scales gracefully with increasing supervision. The instruction-conditioned hyperdimensional representation reduces the reliance on large labeled datasets while preserving robust and consistent sensing parameter recommendation. Such behavior is well aligned with industrial inspection scenarios, where comprehensive data collection across all operating conditions is often costly or impractical.

\subsection{Ablation Study}

\begin{table*}[]
\centering
\caption{Modality ablation results for scanning parameter recommendation. We evaluate three configurations to analyze the contribution of each modality: Observation-only, Instruction-only, and Observation+Instruction.}
\label{tab:ablation study}
\resizebox{\linewidth}{!}{
\setlength{\tabcolsep}{3pt}
\renewcommand{\arraystretch}{1.15}
\begin{tabular}{l|
ccc ccc ccc ccc ccc|ccc}
\toprule
\multirow{2}{*}{\textbf{Setting}} 
& \multicolumn{3}{c}{\textbf{Sample Freq}} 
& \multicolumn{3}{c}{\textbf{Measurement Range X}} 
& \multicolumn{3}{c}{\textbf{Exposure Time}} 
& \multicolumn{3}{c}{\textbf{CMOS Dynamic Range}} 
& \multicolumn{3}{c}{\textbf{Light Intensity Range}} 
& \multicolumn{3}{c}{\textbf{Average}} \\
\cmidrule(lr){2-16}\cmidrule(lr){17-19}
& Exact & Win@1 & F1
& Exact & Win@1 & F1
& Exact & Win@1 & F1
& Exact & Win@1 & F1
& Exact & Win@1 & F1
& Exact & Win@1 & F1 \\
\midrule
\textbf{Observation Only} 
& 30.0$\pm$2.4 & 73.8$\pm$5.9 & 29.0$\pm$1.7 
& 70.2$\pm$3.9 & - & 72.7$\pm$2.3 
& 66.0$\pm$4.5 & 99.7$\pm$0.7 & 67.5$\pm$5.9 
& 73.4$\pm$3.1 & 100.0$\pm$0.0 & 75.9$\pm$3.1 
& 100.0$\pm$0.0 & 100.0$\pm$0.0 & 100.0$\pm$0.0 
& 67.9$\pm$2.8 & 93.4$\pm$1.5 & 69.0$\pm$2.6 \\
\textbf{Instruction Only} 
& 91.2$\pm$1.5 & 94.0$\pm$1.9 & 91.4$\pm$1.2 
& 94.8$\pm$0.8 & - & 93.9$\pm$0.9 
& 85.7$\pm$2.3 & 98.5$\pm$0.7 & 86.0$\pm$2.4 
& 84.3$\pm$0.7 & 96.6$\pm$0.8 & 79.6$\pm$0.6 
& 93.6$\pm$1.6 & 95.7$\pm$0.6 & 90.9$\pm$1.5 
& 89.9$\pm$1.4 & 96.2$\pm$0.6 & 88.4$\pm$1.3 \\
\textbf{Observation+Instruction} 
& 90.6$\pm$2.6 & 94.3$\pm$1.4 & 90.5$\pm$2.7 
& 96.5$\pm$1.9 & - & 97.1$\pm$1.5 
& 90.6$\pm$1.9 & 99.7$\pm$0.5 & 90.3$\pm$1.9 
& 86.4$\pm$0.7 & 99.9$\pm$0.2 & 81.5$\pm$1.7 
& 99.3$\pm$0.7 & 100.0$\pm$0.0 & 99.4$\pm$0.5 
& 92.7$\pm$1.6 & 98.5$\pm$0.4 & 91.8$\pm$1.7 \\
\bottomrule
\end{tabular}
}
\end{table*}

We conduct a modality ablation study to analyze the contribution of visual observations and inspection instructions under cross-split generalization settings. Specifically, we evaluate three configurations: Observation-only, Instruction-only, and Observation+Instruction. All models are trained and evaluated under the same position, rotation, and lighting splits, and performance is reported using the same evaluation protocol as in previous experiments.

Table~\ref{tab:ablation study} summarizes the ablation results across all scanning parameters. The Observation-only configuration exhibits limited performance, with an average Exact Accuracy of $67.9\%$ and F1 score of $69.0\%$. While this setting achieves high Win@1 Accuracy for appearance-sensitive parameters such as light intensity range, its performance remains substantially lower on intent-driven parameters, including sampling frequency and measurement range. These results indicate that visual observations alone are insufficient to reliably infer scanning configurations that depend on inspection intent.

The Instruction-only configuration shows a marked improvement over Observation-only, achieving an average Exact Accuracy of $89.9\%$ and F1 score of $88.4\%$. This setting performs strongly on parameters that are closely tied to instruction semantics, such as sampling frequency and measurement range. However, its performance degrades on appearance-sensitive parameters, particularly exposure time and CMOS dynamic range, where the absence of visual cues limits the model’s ability to resolve surface- and illumination-dependent variations.

The Observation+Instruction configuration achieves the best overall performance, with an average Exact Accuracy of $92.7\%$, Win@1 Accuracy of $98.5\%$, and F1 score of $91.8\%$. Compared to Instruction-only, incorporating visual observations yields consistent improvements on appearance-sensitive parameters, including exposure time and CMOS dynamic range, while preserving high accuracy on intent-driven parameters. This combination enables balanced performance across all scanning parameters and reduces error accumulation under cross-split conditions.

Overall, the ablation results demonstrate that neither instruction semantics nor visual observations alone are sufficient for reliable scanning parameter recommendation. Instructions specify the inspection intent and constrain the admissible parameter space, while visual observations provide essential cues for resolving appearance-dependent parameters. Their combination enables ScanHD to achieve consistently high accuracy under spatial and illumination shifts, supporting the effectiveness of instruction-conditioned multimodal reasoning.

\subsection{Inference Efficiency and Deployment Cost}

We analyze the inference efficiency of ScanHD in comparison with representative multimodal baselines to assess their practical deployment cost. All experiments are conducted on a single NVIDIA RTX~4090 GPU using identical software configurations and batch sizes. Inference latency is measured as the average time required to produce a complete scanning parameter prediction for one input instance.

\begin{figure}[]
  \centering
  \includegraphics[width=\linewidth]{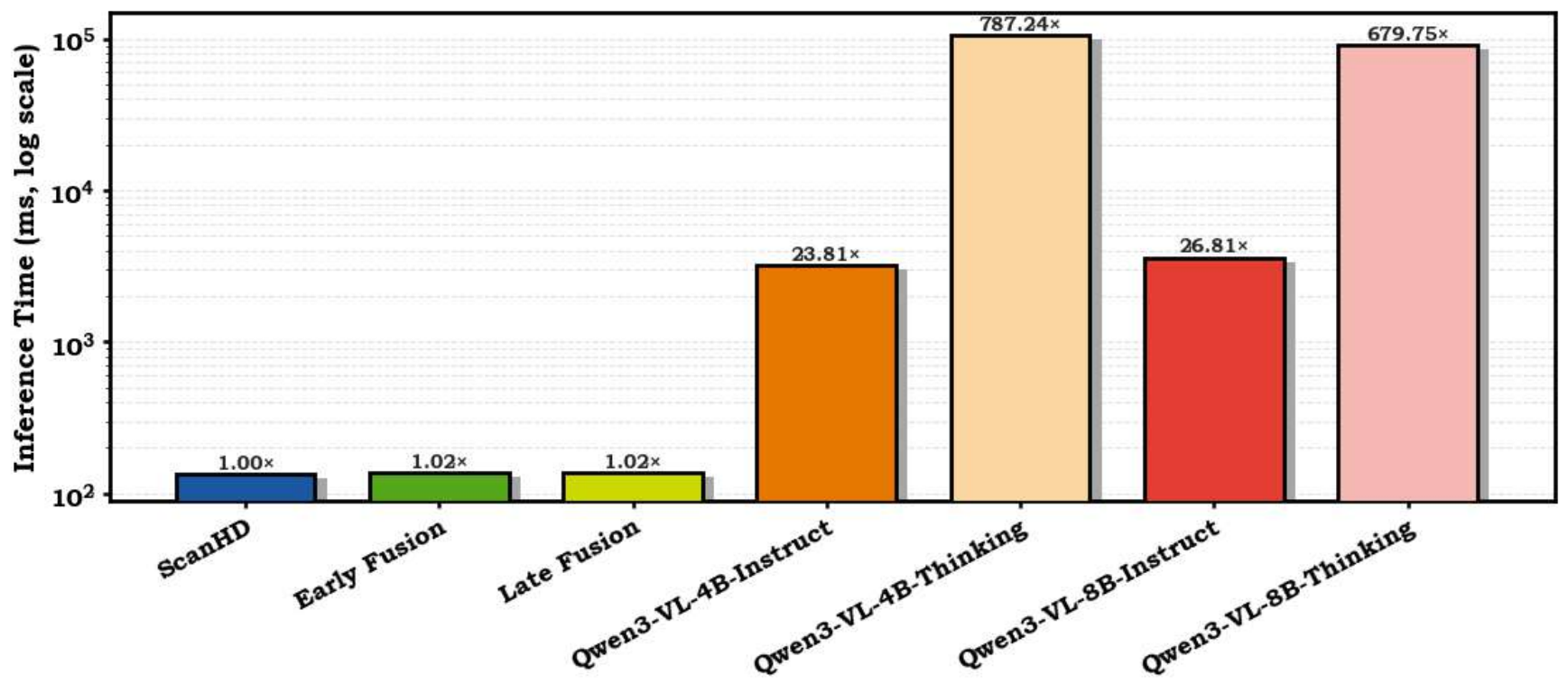}
  \caption{Comparison of inference latency across different methods.}
  \label{exp: inference time}
\end{figure}

As shown in Figure~\ref{exp: inference time}, ScanHD achieves the lowest inference latency among all evaluated methods. Its inference cost remains comparable to lightweight multimodal fusion baselines, with Early Fusion and Late Fusion incurring only marginal overhead relative to ScanHD. This observation confirms that the instruction-conditioned hyperdimensional inference in ScanHD introduces minimal computational burden beyond basic feature fusion.

In contrast, multimodal large language models exhibit substantially higher inference latency. Qwen3-VL-4B-Instruct requires over an order of magnitude more time than ScanHD, while Qwen3-VL-8B-Instruct further increases the latency. The Thinking variants incur a dramatically higher cost, with inference time exceeding ScanHD by several hundred times. The observed efficiency gap has important implications for embodied industrial inspection. Scanning parameter recommendation is often executed in the control loop of robotic systems and must satisfy strict latency and predictability requirements. The compact associative inference of ScanHD enables fast and deterministic execution, whereas the high and variable latency of large multimodal language models limits their suitability for real-time or resource-constrained deployment.

\section{Conclusion and Future Work} \label{Conclusion}

This paper studied instruction-conditioned sensing parameter reasoning as a fundamental problem in embodied industrial inspection. Unlike conventional robotic perception pipelines that assume fixed sensing configurations, we treated the sensor as an adaptive component whose operating parameters should be inferred jointly from inspection intent expressed in natural language and observable scene characteristics. To enable systematic investigation, we introduced Instruct-Obs2Param, a real-world multimodal dataset collected on an industrial robotic scanning platform equipped with a UR3 manipulator and a Keyence LJ-X8200 laser profiler. The dataset captures diverse objects, inspection tasks, viewpoints, and illumination conditions, explicitly modeling the interaction between natural-language instructions, visual appearance, and discrete scanner parameter choices. By grounding parameter labels in expert-calibrated operating regimes and decoupling inspection intent from appearance variation, the dataset provides a practical and realistic benchmark for task-aware sensing adaptation. Building on this dataset, we proposed ScanHD, a hyperdimensional computing–based framework for robust and interpretable sensing parameter recommendation. ScanHD binds visual observations and inspection instructions into a symbolic hyperdimensional representation and performs parameter-wise associative reasoning using compact memories. Extensive experiments on the Instruct-Obs2Param benchmark demonstrated that ScanHD consistently outperforms rule-based heuristics, conventional multimodal learning methods, and multimodal large language models across multiple scanner parameters. In addition, ScanHD exhibits strong robustness to viewpoint and illumination changes while maintaining low computational overhead, making it well suited for real-time deployment in industrial inspection systems. These results highlight the limitations of purely semantic or end-to-end prediction approaches for sensing configuration and validate the effectiveness of hyperdimensional representations for instruction-conditioned sensing reasoning.

This work opens several promising avenues for future research. First, while ScanHD focuses on pre-scan parameter recommendation, future systems could incorporate closed-loop sensing, where scanner parameters are adjusted online based on intermediate scan quality, uncertainty estimates, or detected anomalies. Such integration would enable unified reasoning over how to sense and where to sense within active inspection pipelines. Second, extending the framework beyond discretized operating regimes to support continuous or hybrid discrete–continuous parameter spaces would allow finer-grained control for precision metrology and advanced inspection tasks. Third, although this study is grounded in a laser profiling system, the proposed formulation is sensor-agnostic; future work could generalize instruction-conditioned parameter reasoning to other sensing modalities and multi-sensor inspection platforms. Finally, future work may extend sensing parameter reasoning toward a complete embodied inspection pipeline, in which parameter recommendation, physical scan execution, and downstream inspection tasks are jointly considered. By explicitly linking sensor configuration to scan quality and task-level performance, such as defect detection or dimensional verification, inspection systems could reason holistically across the entire sensing–analysis process and enable scalable, adaptive deployment in real-world manufacturing environments.


\section*{CRediT authorship contribution statement}
\textbf{Zhiling Chen:} Methodology, Software, Data Curation, Formal Analysis, Writing - Original Draft, Writing – Review \& Editing.
\textbf{David Gorsich:} Writing – Review \& Editing, Supervision, Resources, Project administration, Conceptualization.
\textbf{Matthew P. Castanier:} Writing – Review \& Editing, Supervision, Resources, Project administration, Conceptualization.
\textbf{Yang Zhang:} Data Curation, Writing – Review \& Editing.
\textbf{Jiong Tang:} Supervision, Writing – Review \& Editing.
\textbf{Farhad Imani:} Conceptualization, Supervision, Writing – Review \& Editing, Funding Acquisition. 

\section*{Declaration of Competing Interest}
The authors declare that they have no known competing financial interests or personal relationships that could have appeared to influence the work reported in this paper.

\section*{Acknowledgments}
This work was supported under Cooperative Agreement W56HZV-21-2-0001 with the US Army DEVCOM Ground Vehicle Systems Center (GVSC), through the Virtual Prototyping of Autonomy Enabled Ground Systems (VIPR-GS) program, and by the National Science Foundation, United States (Grant No. 2434519).

DISTRIBUTION STATEMENT A. Approved for public release; distribution is unlimited. OPSEC10423

\section*{Disclaimer}
Reference herein to any specific commercial company, product, process, or service by trade name, trademark, manufacturer, or otherwise, does not necessarily constitute or imply its endorsement, recommendation, or favoring by the United States Government or the Department of the Army (DoA). The opinions of the authors expressed herein do not necessarily state or reflect those of the United States Government or the DoA and shall not be used for advertising or product endorsement purposes.

\balance
 \bibliographystyle{elsarticle-num-names} 
 \bibliography{cas-refs}

\end{document}